\documentclass{article}
\usepackage{ijcai23}
\usepackage{times}
\usepackage{soul}
\usepackage{url}
\usepackage[hidelinks]{hyperref}
\usepackage[utf8]{inputenc}
\usepackage[small]{caption}
\usepackage{graphicx}
\usepackage{amsmath}
\usepackage{amsthm}
\usepackage{booktabs}
\usepackage{algorithm}
\usepackage[noend]{algpseudocode}
\usepackage[switch]{lineno}
\usepackage{todo}
\usepackage{nicefrac}
\usepackage{bm}
\usepackage{multicol}
\usepackage{multirow}
\usepackage[nameinlink]{cleveref}
\usepackage{subcaption}
\usepackage{tikz}
\makeatletter
\newcommand*{\T}{{\mathpalette\@transpose{}} }
\newcommand*{\@transpose}[2]{\raisebox{\depth}{$\m@th#1\intercal$}}
\makeatother
\usetikzlibrary{positioning,shapes,bayesnet,calc,backgrounds}

\makeatletter
\newif\if@showgrid@grid
\newif\if@showgrid@left
\newif\if@showgrid@right
\newif\if@showgrid@below
\newif\if@showgrid@above
\tikzset{%
  every show grid/.style={},
  show grid/.style={execute at end picture={\@showgrid{grid=true,#1}}},%
  show grid/.default={true},
  show grid/.cd,
  labels/.style={font={\sffamily\small},help lines},
  xlabels/.style={},
  ylabels/.style={},
  keep bb/.code={\useasboundingbox (current bounding box.south west) rectangle (current bounding box.north west);},
  true/.style={left,below},
  false/.style={left=false,right=false,above=false,below=false,grid=false},
  none/.style={left=false,right=false,above=false,below=false},
  all/.style={left=true,right=true,above=true,below=true},
  grid/.is if=@showgrid@grid,
  left/.is if=@showgrid@left,
  right/.is if=@showgrid@right,
  below/.is if=@showgrid@below,
  above/.is if=@showgrid@above,
  false,
}
\def\@showgrid#1{%
  \begin{scope}[every show grid,show grid/.cd,#1]
    \if@showgrid@grid
      \begin{pgfonlayer}{background}
        \draw [help lines]
        (current bounding box.south west) grid
        (current bounding box.north east);
        \pgfpointxy{1}{1}%
        \edef\xs{\the\pgf@x}%
        \edef\ys{\the\pgf@y}%
        \pgfpointanchor{current bounding box}{south west}
        \edef\xa{\the\pgf@x}%
        \edef\ya{\the\pgf@y}%
        \pgfpointanchor{current bounding box}{north east}
        \edef\xb{\the\pgf@x}%
        \edef\yb{\the\pgf@y}%
        \pgfmathtruncatemacro\xbeg{ceil(\xa/\xs)}
        \pgfmathtruncatemacro\xend{floor(\xb/\xs)}
        \if@showgrid@below
          \foreach \X in {\xbeg,...,\xend} {
            \node [below,show grid/labels,show grid/xlabels] at (\X,\ya) {\X};
          }
        \fi
        \if@showgrid@above
          \foreach \X in {\xbeg,...,\xend} {
            \node [above,show grid/labels,show grid/xlabels] at (\X,\yb) {\X};
          }
        \fi
        \pgfmathtruncatemacro\ybeg{ceil(\ya/\ys)}
        \pgfmathtruncatemacro\yend{floor(\yb/\ys)}
        \if@showgrid@left
          \foreach \Y in {\ybeg,...,\yend} {
            \node [left,show grid/labels,show grid/ylabels] at (\xa,\Y) {\Y};
          }
        \fi
        \if@showgrid@right
          \foreach \Y in {\ybeg,...,\yend} {
            \node [right,show grid/labels,show grid/ylabels] at (\xb,\Y) {\Y};
          }
        \fi
      \end{pgfonlayer}
      \fi
    \end{scope}
  }
  \makeatother
  \tikzset{every show grid/.style={show grid/keep bb}%
  }%

\newcommand{\mypar}[1]{{\bf #1.}\quad}
\newcommand{\myparquestion}[1]{{\bf #1}\quad}
\newcommand{\disprod}{DiSProD}
\newcommand{\disprodSpace}{DiSProD }
\newcommand{\citeAY}[1]{\citeauthor{#1} [\citeyear{#1}]}

\title{\disprod: Differentiable Symbolic Propagation of Distributions for Planning}

\author{
Palash Chatterjee\and
Ashutosh Chapagain\and
Weizhe Chen\And
Roni Khardon \\
\affiliations
Department of Computer Science, Luddy School of Informatics, Computing, and Engineering, Indiana University, Bloomington, Indiana, USA\\
\emails
\{palchatt, aschap, chenweiz, rkhardon\}@iu.edu
}

\begin{document}
\maketitle
\begin{abstract}
The paper introduces \disprod, an online planner developed for environments with probabilistic transitions in continuous state and action spaces. \disprodSpace builds a symbolic graph that captures the distribution of future trajectories, conditioned on a given policy, using independence assumptions and approximate propagation of distributions. The symbolic graph provides a differentiable representation of the policy's value, enabling efficient gradient-based optimization for long-horizon search. The propagation of approximate distributions can be seen as an aggregation of many trajectories, making it well-suited for dealing with sparse rewards and stochastic environments. An extensive experimental evaluation compares \disprodSpace to state-of-the-art planners in discrete-time planning and real-time control of robotic systems. The proposed method improves over existing planners in handling stochastic environments, sensitivity to search depth, sparsity of rewards, and large action spaces. Additional real-world experiments demonstrate that \disprodSpace can control ground vehicles and surface vessels to successfully navigate around obstacles.
\end{abstract}
% !TEX root =  main.tex
\section{Introduction}%
\label{sec:introduction}
Planning is one of the key problems in artificial intelligence~(AI), as it enables intelligent agents to make informed decisions and achieve their objectives in complex 
and dynamic 
environments.
This is especially important when the environment is inherently stochastic or when the dynamics model used by the planning algorithm is imperfect.
%Planning algorithms should represent and reason about uncertainty to make robust and effective decisions.
As a result, research on planning with stochastic transitions has been multifaceted, encompassing symbolic task-level planning in discrete spaces \cite{kolobov2012reverse,keller2013trial,cui2019stochastic}, 
robotic 
motion planning in continuous spaces \cite{kurniawati2011motion,van2012motion,agha2014firm}, integrated task and motion planning \cite{kaelbling2013integrated,vega2018admissible,garrett2021integrated} and model-based reinforcement learning 
%leveraging probabilistic dynamics 
\cite{chua2018deep,hafner2019dream,curi2020efficient}.

Markov Decision Processes~(MDPs) provide the theoretical foundation for planning under uncertainty, but scalability remains a challenge, and approximation is often necessary.
Many approaches have been proposed in the literature, including searching in action space, searching in state space, approaches using Monte-Carlo simulation, and approaches using differentiation.
Several gradient-based planners have been proposed, but they are not easily usable as domain independent planners due to scalability \cite{deisenroth2011pilco}, or restriction to  deterministic environments \cite{wu2020scalable}, or in the Reinforcement Learning (RL) context, where, in most works the success of the planner depends on the quality of the learned domain-specific value function \cite{tamar2016value,hafner2019dream}.

The paper fills this gap by introducing a novel domain-independent online planner using differentiable probabilistic transition models.
Our work is related to prior algorithms for trajectory optimization, distribution propagation, and differential dynamic programming \cite{chua2018deep,williams2017information,deisenroth2013gaussian,tassa2012synthesis,lenz2015deepmpc}.
However, we introduce a novel symbolic approximation and propagation scheme generalizing work on AI planning that facilitate robustness, better approximation, and optimization \cite{cui2019stochastic}.

The main contribution of this work is in designing \disprodSpace (\textbf{Di}fferentiable \textbf{S}ymbolic \textbf{Pro}pagation of \textbf{D}istributions), an online planner for environments with probabilistic transitions,
in continuous or hybrid spaces.
The core idea is to create a symbolic graph that encodes the distribution of future trajectories conditioned on a given policy.
The resulting symbolic graph provides an \emph{analytically differentiable representation} of the policy's value, allowing efficient gradient-based optimization of the action variables for long-horizon search.
While distributions over trajectories are too complex to be captured exactly, \disprodSpace uses Taylor's approximation and an independence assumption to facilitate a symbolic propagation of product distributions over future state and reward variables.
The approximate distribution propagation can be viewed as an efficient symbolic aggregation of many trajectories, which differs from sampling algorithms 
that aggregate the results of many individual trajectories.
This approach reduces variance in estimates in stochastic environments and facilitates planning with sparse rewards.

Extensive quantitative experiments are conducted to compare \disprodSpace with state-of-the-art planners in discrete-time planning in OpenAI Gym environments \cite{brockman2016openai}  and continuous-time control of simulated robotic systems.
The results show that \disprodSpace outperforms existing planners in dealing with stochastic environments, sensitivity to search depth, sparsity of rewards, and large action spaces.
Furthermore, we use \disprodSpace with an approximate transition model to control two real-world robotic systems demonstrating that it can successfully control ground vehicles and surface vessels to navigate around obstacles. 

Due to space constraints, some details are omitted from the paper. 
The full paper as well as code to reproduce the experiments and videos from physical experiments are available at \url{https://pecey.github.io/DiSProD}.
% !TEX root =  main.tex
\section{Related Work}\label{sec:related_work}
Planning in continuous spaces has been studied in several sub-fields.
A key distinction in planning methods is between \emph{offline planners}, which compute a complete solution and then apply it, and \emph{online planners} or Model Predictive Control~(MPC)~\cite{borrelli2017predictive}, in which at every time-step, optimization is carried out over a finite horizon and only the first action from the solution is executed.
The online nature of MPC provides some robustness to unexpected outcomes at the cost of increased computation time during action selection.
Another important distinction is planning in \emph{state space} (or configuration space) ~\cite{lavalle2006planning} versus planning in \emph{action space}.
The former seeks an optimal sequence of states, leaving action execution to a low-level controller, while the latter produces executable actions directly.
\disprodSpace is an \emph{online action planner} but it can also produce a sequence of states as a byproduct, as we discuss in \Cref{sec: exp_with_robotics_systems}.

Within online action planners, \disprodSpace
is related to two lines of work using differentiable transition models.
First, our approach builds on the 
SOGBOFA algorithm
\cite{cui2018stochastic,cui2019stochastic}, 
which was developed for discrete task-level AI planning problems.
However, that work is restricted to binary state and action variables.
In addition, \disprodSpace introduces a new distribution propagation method based on symbolic Taylor expansions.
The second group includes planners using
differentiable transition models in RL and control.
However, deep RL work (e.g., \cite{heess2015learning,depeweg2016learning}) makes use of learned value functions to aid planning performance and cannot plan in a new model without training first, and many approaches (e.g., \cite{wu2020scalable}) use deterministic transition models.
In addition, most approaches \cite{hafner2019dream,Bueno2019deep}
optimize over individual trajectories and do not propagate distributions over trajectories as in DiSProD.
In this realm, iLQG \cite{tassa2012synthesis} and the PILCO family \cite{deisenroth2013gaussian,parmas2018pipps,kamthe2018data} are most related to our method.
iLQG linearizes the dynamics, assumes linear Gaussian transitions, as in the Extended Kalman Filter~(EKF), and optimizes over individual trajectories.
PILCO does propagate distributions analytically, albeit with the restricted Gaussian process~(GP) dynamics and the Gaussian kernel.
\citeAY{gal2016improving} replace the GP in PILCO with Bayesian neural networks, which cannot propagate distributions analytically and hence requires particle-based planning.
From this perspective, DiSProD can be seen as a generalization of PILCO and iLQG, which uses differentiation over approximate symbolic propagation of distribution.

Our work is also related to sampling-based planners
Model Predictive Path Integral~(MPPI) and Cross-Entropy Method~(CEM) \cite{kobilarov2012cross,williams2017information,chua2018deep,wagener2019online,mohamed2022autonomous}.
These algorithms sample a set of trajectories around a nominal trajectory using a fixed sampling distribution, and update the nominal trajectory based on the ``goodness'' of the trajectories to bias sampling in subsequent iterations.
Similarly, \citeAY{Mania2018} use sampling to estimate numerical gradients which provide related policy updates. 
Our framework uses distribution propagation instead of sampling and it optimizes stochastic policies instead of having a fixed sampling distribution.

Finally, we note that our approach can flexibly handle both discrete and continuous variables.
In contrast, other methods for planning in hybrid spaces typically restrict the transitions, for example, to piecewise linear models \cite{li2005lazy,zamani2012symbolic,raghavan2017hindsight}.

% !TEX root =  main.tex
\section{Algorithms and Methodology}%
\label{sec:alg}
A Markov Decision Process~(MDP) is specified by $\{\mathcal{S}, \mathcal{A}, {T}, \mathcal{R}, \gamma\}$, where $\mathcal{S}$ is the state space, $\mathcal{A}$ is the action space, ${T}$ is the transition function, $\mathcal{R}$ is the one-step reward function, and $\gamma$ is the discount factor. A policy $\pi$ is a mapping from states to actions.
Given a policy $\pi$, the action-value function $\mathcal{Q}^\pi(\bm{s}_t,\bm{a}_t)$ represents the expected discounted total reward that can be obtained from state $\bm{s}_t$ if action $\bm{a}_t$ is chosen, and the policy $\pi$ is followed thereafter, where $\bm{s}_t$ and $\bm{a}_t$ are vectors of state and action variables 
at time $t$.
We focus on open loop probabilistic policies parameterized by $\theta=\{\theta_t\}$ where each $\theta_t$ picks the action at time step $t$ and is further factorized over individual action variables. As in prior work, we approximate  $\mathcal{Q}^{\theta}(\bm{s}_t,\theta_t)=\mathbb{E}[\sum_{i = 0}^{D-1} \gamma^i \mathcal{R}(\bm{s}_{t+i}, \bm{a}_{t+i})]$
where $\bm{a}_t\sim p(\bm{a}_t| \theta_t)$ and $\bm{s}_{t+1}\sim p(\bm{s}_{t+1}|\bm{s}_t, \bm{a}_{t})$,
optimize $\theta$ and  pick the action using $\theta_t$.
In principle the $\mathcal{Q}$-value can be calculated from the distributions over $\{(\bm{s}_{t+i}, \bm{a}_{t+i})\}$ but these distributions are complex.
We approximate these distributions as products of independent distributions over individual state variables.

Our algorithm can be conceptually divided into two parts. 
The first calculates an approximate distribution over future states, rewards and $\mathcal{Q}^\pi(\bm{s}_t,\bm{a}_t)$, conditioned on a given policy. 
A schematic overview of this process is shown in \Cref{fig:hl-diagram}.
The second uses this representation to optimize the policy. These are described in the next two subsections.

%%%%%%%%%%%%%%%%%%%%%%%%%%%%%%%%%%%%%%%%%%%%%%%%%%%%%%%%%%%%%%%
% Analytic Computation Graph for approximating the Q-Function %
%%%%%%%%%%%%%%%%%%%%%%%%%%%%%%%%%%%%%%%%%%%%%%%%%%%%%%%%%%%%%%%
\subsection{Analytic Computation Graph}%
\begin{figure}[t]%
    \centering%
        \includegraphics[width=0.9\linewidth]{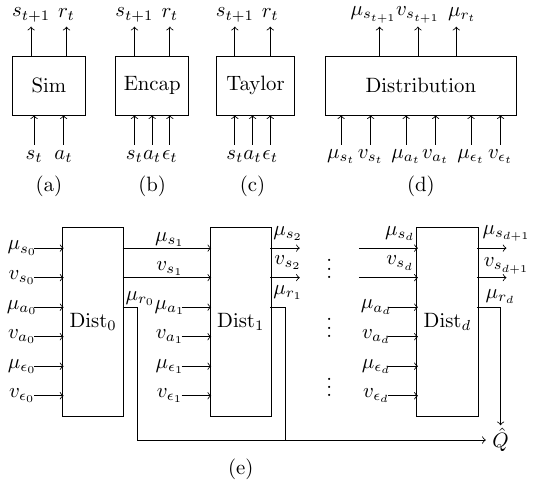}
    \caption{Schematic overview of the idea behind the construction of analytic computation graph. (a) and (b) show the original probabilistic simulator, and its encapsulated variant where noise variables are represented as inputs. Using Taylor's approximation of this variant, we generate a third representation (c) of the same transition function. All these take as input a concrete state, action (and noise) values and compute the next state. The key idea is to combine the approximation with propagation of distributions to yield (d) that takes distributions on states, actions, and noise as input and produces a distribution over next states. Stacking this model up to the desired search depth yields symbolic propagation of distributions (e).}\label{fig:hl-diagram}%
\end{figure}

\label{sec: analytic_graph}
% \mypar{Transition Model}
\paragraph{Transition Model.}
Typically, given $\bm{s}_t$ and $\bm{a}_t$, a simulator $T_{\text{sim}}$, samples from a distribution over $\bm{s}_{t+1}$.  Formally, $\bm{s}_{t+1} \sim T_{\text{sim}}(\bm{s}_t, \bm{a}_t)$.
For DiSProD, we need to \emph{encapsulate} any internal sampling in $T_{\text{sim}}$ as an input, to separate sampling from computation, similar to the reparameterization trick.
\begin{align}
\bm{s}_{t+1} = T(\bm{s}_t, \bm{a}_t, \bm{\epsilon}_t) \mbox{ where } \bm{\epsilon}_t \sim \mathcal{N}(0, I).
\label{eq:encapsulate}
\end{align}
$T_{\text{sim}}$ and its encapsulation are shown in \Cref{fig:hl-diagram}a and 1b.

% \mypar{Representation}
\paragraph{Representation.}
Let $s_{t,k}$ denote the value of a random variable $s_k$ at time $t$. If ${s}_{t,k}$ is binary, then its marginal distribution  
can be captured using 
just its mean $\hat{\mu}_{s,t,k}$
as the variance is implicitly given by $\hat{v}_{s,t,k} = \hat{\mu}_{s,t,k} (1-\hat{\mu}_{s,t,k})$.
If ${s}_{t,k}$ is continuous, its distribution can be summarized using its mean and variance $(\hat{\mu}_{s,t,k},\hat{v}_{s,t,k})$.
Similarly, the distribution over the noise variable $\epsilon_i$ 
and action variable $a_\ell$ are
represented using $(\hat{\mu}_{\epsilon,t,i},\hat{v}_{\epsilon,t,i})$ and $(\hat{\mu}_{a,t,\ell},\hat{v}_{a,t,\ell})$ respectively.
While the use of mean and variance suggests a normal distribution, the proposed construction 
does not assume anything about the form of the distribution.

% \mypar{Independence Assumption}
\paragraph{Independence Assumption.}
To simplify the computations of propagated distributions, our approximation assumes that for all $t$, the distribution over the trajectories given by $p(\bm{s}_t,\bm{a}_t,\bm{\epsilon}_t)$ is a product over independent factors: $p(\bm{s}_t,\bm{a}_t,\bm{\epsilon}_t)=\prod_k p({s}_{t,k}) \prod_\ell p({a}_{t,\ell}) \prod_i p(\epsilon_{t,i})$.

% \mypar{Approximation}
\paragraph{Approximation.}

Our main observation is that the above assumption suffices to propagate approximate distributions and approximate the $\mathcal{Q}$-function.
Specifically, we approximate the transition function using a second order Taylor expansion whose terms are given analytically in a symbolic form.
Let $\bm{z}_t=$ $(\bm{s}_t,\bm{a}_t,\bm{\epsilon}_t)$, 
$\hat{\bm{z}}_t=$ $(\hat{\mu}_{\bm{s}_t},\hat{\mu}_{\bm{a}_t},\hat{\mu}_{\bm{\epsilon}_t})$ and $\bm{s}_{t+1} =$ $T(\bm{z}_t)$.
To simplify the notation, consider the $j$-th state variable 
$s_{t+1,j}=T_j(\bm{s}_t, \bm{a}_t, \bm{\epsilon}_t) = T_j(\bm{z}_t)$ and let $\nabla T_j = \nicefrac{\partial T_j}{\partial \bm{z}_t}$ and $H_j=\nicefrac{\partial^2 T_j}{\partial \bm{z}_t \partial \bm{z}_t^\top}$.
We use Taylor's expansion to approximate the encapsulated model in \Cref{eq:encapsulate} by
\begin{align}
    s_{t+1,j}\leftarrow T_j(\bm{z}_t) &\approx T_j(\hat{\bm{z}_t}) + \nabla T_j^\top (\bm{z}_t-\hat{\bm{z}_t})\nonumber\\
                                 &+ \frac{1}{2}(\bm{z}_t-\hat{\bm{z}_t})^\top H_j (\bm{z}_t-\hat{\bm{z}_t}).
\end{align}
This 
is illustrated in \Cref{fig:hl-diagram}c.
We use this approximation to calculate the mean and variance of $\bm{s}_{t+1}$ via methods of propagating distributions.
Given our independence assumption, the off diagonal covariance terms multiplying $H_j$ are zero, and the expected value of the $s_{t+1, j}$ becomes
\begin{align}
    &\hat{\mu}_{s,t+1,j} =\mathbb{E}\left[s_{t+1, j}\right]\approx{T}_j(\hat{\bm{z}_t})+\frac{1}{2}\Biggl[\sum_k\left(\frac{\partial^2 T_j}{\partial{s}_{t,k}^2}\right)\hat{v}_{s,t,k}\nonumber\\
    &+\sum_\ell\left(\frac{\partial^2{T}_j}{\partial{a}_{t,\ell}^2}\right)\hat{v}_{a,t,\ell}+\sum_{i}\left(\frac{\partial^2{T}_j}{\partial{\epsilon}_{t,i}^2}\right)\hat{v}_{\epsilon,t,i}\Biggr].
\end{align}

A similar approximation for the variance yields $4^\text{th}$ order moments.
To reduce complexity, we use a first order Taylor approximation for the variance resulting in
\begin{align}
    \hat{v}_{s,t+1,j} &\approx \sum_k \biggl(\frac{\partial T_j}{\partial {s}_{t,k}} \biggr)^2 \hat{v}_{s,t,k} + \sum_\ell \biggl(\frac{\partial T_j}{\partial {a}_{t,\ell}} \biggr)^2 \hat{v}_{a,t,\ell} \nonumber \\
                      &+ \sum_i \biggl(\frac{\partial T_j}{\partial {\epsilon}_{t,i}} \biggr)^2 \hat{v}_{\epsilon,t,i}.
\end{align}
To write these concisely 
we collect the first order partials and diagonals of the Hessians into matrices as follows:
\begin{align}
  J^{T}_{\bm{s}_t} &= \biggl[\frac{\partial T}{\partial \bm{s}_{t}^\top}\biggr],
               & \mbox{ i.e.,\ \ \ \ } & [J^{T}_{\bm{s}_t}]_{j,k}= \frac{\partial T_j}{\partial s_{t,k}},
               \\
  \tilde{H}^{T}_{\bm{s}_t} &= \biggl[\frac{\partial^2 T}{\partial \bm{s}_{t}^2}^\top \biggr],
                       & \mbox{ i.e.,\ \ \ \ \ } & [\tilde{H}^{T}_{\bm{s}_t}]_{j,k}= \frac{\partial^2 T_j}{\partial s_{t,k}^2}.
\end{align}
Similarly, we define $J^{T}_{\bm{a}_t},\tilde{H}^{T}_{\bm{a}_t}, J^{T}_{\bm{\epsilon}_t}, \text{ and } \tilde{H}^{T}_{\bm{\epsilon}_t}$ for action and noise variables, respectively. We also define $\tilde{H}^{\mathcal{R}}_{\bm{s}_t}$, $\tilde{H}^{\mathcal{R}}_{\bm{a}_t}$, \text{ and } $\tilde{H}^{\mathcal{R}}_{\bm{\epsilon}_t}$ to be the second order partials of the reward function $\mathcal{R}$ with respect to the state, action and noise variables. 
We can now write the vector form of the equations as follows:
\begin{align}
  % Expectation
    \hat{\mu}_{\bm{s}_{t+1}} &\approx T(\hat{z_t}) + \frac{1}{2}\biggl[ \tilde{H}^{T}_{\bm{s}_t} \hat{v}_{\bm{s}_t} + \tilde{H}^{T}_{\bm{a}_t} \hat{v}_{\bm{a}_t} + \tilde{H}^{T}_{\bm{\epsilon}_t} \hat{v}_{\bm{\epsilon}_t} \biggr].
  \label{eq:expectation_vector_form}
  \\
  % Variance
  \hat{v}_{\bm{s}_{t+1}}  &\approx (J^{T}_{\bm{s}_t} \odot J^{T}_{\bm{s}_t}) \hat{v}_{\bm{s}_t} + (J^{T}_{\bm{a}_t} \odot J^{T}_{\bm{a}_t}) \hat{v}_{\bm{a}_t}\notag\\
                          &+ (J^{T}_{\bm{\epsilon}_t} \odot J^{T}_{\bm{\epsilon}_t}) \hat{v}_{\bm{\epsilon}_t}.
  \label{eq:variance_vector_form}
  \\
  % Expectation of reward
    \hat{\mu}_{r_{t}} &\approx \mathcal{R}(\hat{z_t}) + \frac{1}{2}\biggl[ \tilde{H}^{\mathcal{R}}_{\bm{s}_t} \hat{v}_{\bm{s}_t} + \tilde{H}^{\mathcal{R}}_{\bm{a}_t} \hat{v}_{\bm{a}_t} + \tilde{H}^{\mathcal{R}}_{\bm{\epsilon}_t} \hat{v}_{\bm{\epsilon}_t} \biggr].
    \label{eq:reward_expectation_vector_form}
\end{align}
In this paper, we do not model the variance of the reward function. 
However, this can be easily done by analogy with \Cref{eq:variance_vector_form} which will facilitate risk sensitive optimization, for example,
conditional value at risk \cite{ChowGJP17}.
The computations of \Cref{eq:expectation_vector_form,eq:variance_vector_form,eq:reward_expectation_vector_form} are illustrated 
in \Cref{fig:hl-diagram}d.
Stacking these computations over multiple time steps gives us an approximation of the distribution over future states, captured analytically as a computation graph, as shown in \Cref{fig:hl-diagram}e.
The $\mathcal{Q}$-function is then approximated as 
\begin{align}
\label{eq:Qapprox}
\hat{\mathcal{Q}}(\bm{s}_t, \hat{\mu}_{\bm{a}}, \hat{v}_{\bm{a}}) = \sum_{i=0}^{D-1} \gamma^i \hat{\mu}_{r_{t+i}}
\end{align}
where we use $\gamma = 1$ in our experiments.
Note that the computation graph propagates distributions and does not sample trajectories.
The computation only requires the mean $(\hat{\mu}_\epsilon=0)$ and variance $(\hat{v}_\epsilon=1)$ of the noise which are known in advance and are absorbed as constants in the graph.

Our construction is very general -- we only require access to $T(\bm{s}_t,\bm{a}_t,\bm{\epsilon}_t)$, $\mathcal{R}$ and analytic computation of their partial derivatives.
In practice, $T$ and $\mathcal{R}$ can have non-differentiable components which can be mitigated by approximating the non-smooth functions with their smoother alternatives.
\subsection{Optimization Algorithm}
Thanks to the symbolic representation, once the computation graph is built, it can be reused for all steps  of problem solving 
as long as the reward function does not change.
The optimization is conceptually simple, initializing all action variables for all time steps and performing gradient based search. 
However, some important details are discussed next.

% \mypar{Multiple Restarts and Loss function}
\paragraph{Multiple Restarts and Loss function.}
Following \citeAY{cui2019stochastic}, we perform gradient search with multiple restarts.
While they performed restarts sequentially, we take advantage of the structure of the computation to perform the restarts in a vectorized form.
Recall that each restart is represented by $(\hat{\mu}_{\bm{a}}, \hat{v}_{\bm{a}})$ as in \Cref{eq:Qapprox}. Using a superscript $k$ to represent independent restarts we can write the loss function (negating $\mathcal{Q}$) as 
loss = - $\sum_{\text{restart }k} \hat{\mathcal{Q}}(\bm{s_t}, \hat{\mu}_{\bm{a}}^k, \hat{v}_{\bm{a}}^k)$.
Now to evaluate the loss
we generate a matrix where every row represents a possible policy
$(\hat{\mu}_{\bm{a}}^k, \hat{v}_{\bm{a}}^k)$ and evaluate $\hat{\mathcal{Q}}$ on this matrix. 
Therefore, we benefit from evaluating all restarts as a ``batch of examples" relative to the computation graph.
Since we are optimizing with respect to the input matrix (and not computation graph weights), the loss decomposes into a sum of independent functions and gradient search effectively optimizes all restarts simultaneously.

\begin{algorithm}[tb]
  \caption{One Step of DiSProD.}%
  \label{alg:opt}
  \hspace*{\algorithmicindent} \textbf{Input}: state
  \begin{algorithmic}[1]
    \State initialize actions (for all restarts)
    \State build computation graph till depth $D$
    \While{actions have not converged}
    %\State evaluate the computation graph
    \State loss = - $\sum_{\text{restart }k} \hat{\mathcal{Q}}(\bm{s_t}, \hat{\mu}_{\bm{a}}^k, \hat{v}_{\bm{a}}^k)$
    \State loss.backward()
    % \State take a gradient step on actions
    % \State actions $\leftarrow$ projection(actions)
    \State actions  $\leftarrow$ safe-projected-gradient-update(actions)
    \EndWhile
    \State save action-means $\hat{\mu}^{k^*}_{\bm{a}_{t+1:t+D}}$ from the best restart $k^*$\;
    \State \textbf{return} action $\sim \mathcal{N}(\hat{\mu}^{k^*}_{\bm{a}_{t}}, \hat{v}^{k^*}_{\bm{a}_{t}})$ \;
    % \State \textbf{return} action
\end{algorithmic}
\end{algorithm}

% \mypar{Initialization of Actions}
\paragraph{Initialization of Actions.}
For discrete action variables, following \citeAY{cui2019stochastic}, we initialize actions 
parameters for time step $t=0$ to binary values and for steps $t>0$, we initialize the marginals to random values in $[0,1]$.
For continuous action variables, we must constrain actions to be in a valid interval.
For each action variable $a_\ell$, the policy is given by a   
uniform distribution, initializing $\hat{\mu}_{a,\ell} \sim \mathcal{U}(a_{\ell,\text{min}}, a_{\ell,\text{max}})$ and the variance of a uniform distribution centered around $\hat{\mu}_{a,\ell}$, i.e., $\hat{v}_{a,\ell} = \nicefrac{\min(a_{\ell,\text{max}} - \hat{\mu}_{a,\ell}, \hat{\mu}_{a,\ell} - a_{\ell,\text{min}})^2}{12}$.

% \mypar{Policy Updates}
\paragraph{Policy Updates.}
We use 
Adam \cite{kingma2015adam} to optimize the action variables over all restarts simultaneously.
We make at most $K=10$ updates and stop the search early 
when $||\hat{\mu}^{\text{new}}_{\bm{a}} - \hat{\mu}^{\text{old}}_{\bm{a}}||_\infty \le 0.1$ and $||\hat{v}^{\text{new}}_{\bm{a}} - \hat{v}^{\text{old}}_{\bm{a}}||_\infty \le 0.01$ for normalized action ranges.
Gradients are used to update both $\hat{\mu}_a$ and $\hat{v}_a$, which implies that we search over a stochastic policy.
This can be important for success in sparse-reward scenarios
where a stochastic policy effectively broadens the search when needed, while also allowing a nearly deterministic choice of actions at convergence.

% \mypar{Safe Projected-Gradient Update}
\paragraph{Safe Projected-Gradient Update.}
The gradient-based updates have to satisfy two constraints.
The first is the need to constrain variables into valid intervals.
Similar to \cite{tassa2014control,cui2019stochastic},
we use the standard projected gradient descent to restrict $\hat{\mu}_{\bm{a}}$ between ${\bm{a}}_\text{min}$ and $\bm{a}_\text{max}$, while $\hat{v}_{\bm{a}}$ is constrained to $\min(\nicefrac{1}{12}, \nicefrac{\min(\hat{\mu}_{\bm{a}} - \bm{a}_\text{min}, \bm{a}_\text{max} - \hat{\mu}_{\bm{a}})^2}{12})$ which is the variance of the largest legal uniform distribution centered around $\hat{\mu}_{\bm{a}}$. 
If the gradient step pushes $\hat{\mu}_{\bm{a}}$ or $\hat{v}_{\bm{a}}$ outside the valid region, it is clipped to the boundary.
Finally, we only take a gradient step on the restart if it improves the $\mathcal{Q}$-value and otherwise we maintain the previous value.
This safe update requires an additional evaluation of the computation graph to check for the $\mathcal{Q}$-value improvement and it increases runtime, but it ensures that 
$\hat{\mathcal{Q}}$
is monotonically increasing. 

\paragraph{Saving Actions.}
% \mypar{Saving Actions}
Since the gradients are back-propagated through the entire computation graph, action variables at all depths are adjusted according to the corresponding gradients. However, only the action at depth $d=0$ is executed. %Formally, at time $t$, we choose the action $a_{t,0}$ for state $s_t$, while $a_{t,1:D}$ are not used.
Formally, at time $t$, we sample an action using 
$\hat{\mu}_{\bm{a}_{t}}$ and $\hat{v}_{\bm{a}_{t}}$
while 
$\hat{\mu}_{\bm{a}_{t+1:t+D}}$ and $\hat{v}_{\bm{a}_{t+1:t+D}}$ 
are not used.
These updated actions can be used to initialize the action variables when planning at state $s_{t+1}$.
The same idea has been used before in MPPI \cite{williams2017information,wagener2019online}.
Note that MPPI also uses multiple samples, but these samples all contribute to the update of a single action sequence, which can potentially harm the search~\cite{lambert2021stein,barcelos2021dual}.
This is different in \disprodSpace where each restart is an independent search.
To allow reuse of old search but add diversity, we initialize one restart using 
saved action mean,
and initialize all other restarts randomly.

\paragraph{Overall Optimization Algorithm.}
% \mypar{Overall Optimization Algorithm}
These steps are summarized in \Cref{alg:opt},
where search depth $(D)$, the initial step-size for $\hat{\mu}_{\bm{a}}$ and $\hat{v}_{\bm{a}}$ $(lr_{\mu} \text{ and } lr_v)$, and number of restarts serve as hyper-parameters. 
When all restarts have converged or $K$ gradient steps have been performed, we choose the \emph{best restart} with the maximum $\mathcal{Q}$-value, breaking ties randomly.
%Finally, mean of the first action from the best restart is selected.
Finally, we sample an action using the mean and variance of the first action distribution of the best restart.
\subsection{Discussion}
\disprodSpace propagates approximate distributions over trajectories using moment matching on individual variables, assuming independence between variables. As shown by \citeAY{cui2018stochastic}, in the binary case, this is equivalent to belief propagation, which works well empirically despite lack of formal guarantees. \disprodSpace offers a different trade-off from sampling-based methods which are exact in the limit of infinite samples but are sensitive to variance in estimates with limited samples. In this sense, our computation graph provides a stable if biased estimate of the $\mathcal{Q}$-function. 
With deterministic transitions and policy, our computation is exact and gradient descent can potentially find the optimal policy. For stochastic transitions, the computation is approximate and we use a loose stopping criterion 
to reduce run time.

\section{Experiments}%
\label{sec:exp}
We experiment with \disprodSpace on a variety of deterministic and stochastic environments and evaluate its robustness to long horizon planning, its ability to work with sparse rewards, and its performance in high-dimensional action spaces.
For this purpose, we conduct extensive experiments on simulated and real-world environments with increasing complexity.

We use three OpenAI Gym environments (Cart Pole, Mountain Car, and Pendulum) to evaluate the \emph{robustness} of the compared planners in terms of stochasticity, planning horizon, and reward sparsity.
The original OpenAI Gym environments are deterministic.
We therefore
enhance these environments by explicitly adding noise into the model and using it as a part of the dynamics $T(\bm{s},\bm{a},\bm{\epsilon})$. 
Further details are in the full paper. For the discussion, it suffices to note that we parameterize the amount of noise using a scalar $\alpha$ and can therefore evaluate planners' performance as a function of stochasticity in the dynamics.

In addition, we developed a new Gym environment that models a simplified vehicle dynamics, elaborating over the Dubins' car model \cite{dubins1957curves}. In particular, the agent can control the change in linear $(\Delta v_t)$ and angular velocity $(\Delta \omega_t)$ instead of controlling the velocities directly, and the maximum change $(\overline{\Delta{v}}, \overline{\Delta{\omega}})$ is limited to a small value. This model is still inaccurate because it ignores friction, inertia, actuation noise and localization noise, but it provides an acceptable approximation. We then use this model to plan in a physics simulator with asynchronous execution. Specifically, the planner is used to control a TurtleBot in Gazebo simulation via a Robot Operating System~(ROS) interface.

Finally, to demonstrate robustness and applicability, \disprodSpace is used to control two physical robot systems: an Unmanned Ground Vehicle (Jackal) and an Unmanned Surface Vessel (Heron) using the aforementioned model.

\begin{figure*}[th]%
  \centering
  % Row 1
  \begin{subfigure}[b]{0.24\textwidth}
         \centering
         \includegraphics[width=\textwidth]{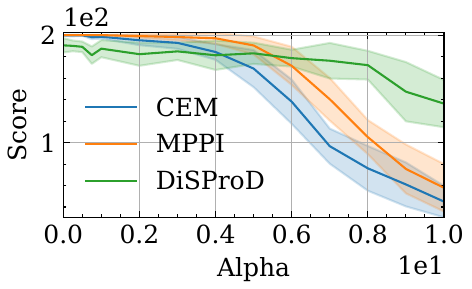}
         \caption{\scriptsize{CCP $(D=25)$}}
         \label{fig:exp_noise_cartpole}
    \end{subfigure}
    \hfill
     \begin{subfigure}[b]{0.24\textwidth}
         \centering
         \includegraphics[width=\textwidth]{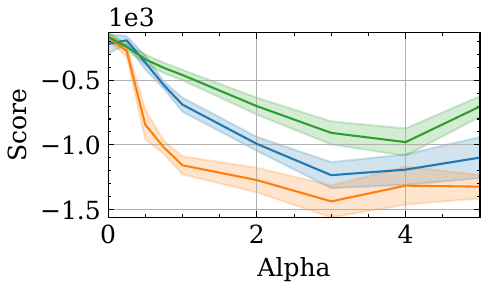}
        \caption{\scriptsize{P $(D=25)$}}
         \label{fig:exp_noise_pendulum}
    \end{subfigure}
    \hfill
     \begin{subfigure}[b]{0.24\textwidth}
         \centering
         \includegraphics[width=\textwidth]{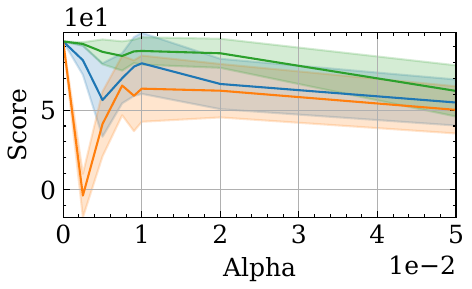}
         \caption{\scriptsize{CMC $(D=100)$}}
         \label{fig:exp_noise_mountain_car}
    \end{subfigure}
    \hfill
     \begin{subfigure}[b]{0.24\textwidth}
         \centering
         \includegraphics[width=\textwidth]{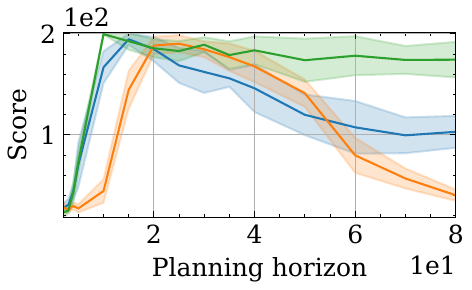}
         \caption{\scriptsize{CCP $(\alpha=5)$}}
         \label{fig:exp_depth_cartpole_noisy}
    \end{subfigure}
    \hfill
    % Row 2 
    \begin{subfigure}[b]{0.24\textwidth}
         \centering
         \includegraphics[width=\textwidth]{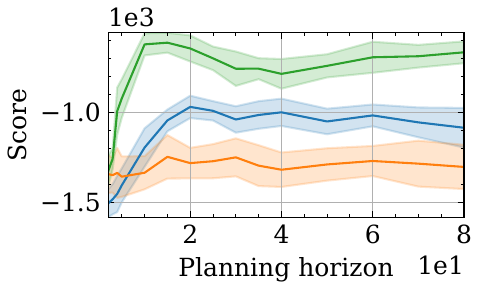}
         \caption{\scriptsize{P $(\alpha=2)$}}
         \label{fig:exp_depth_pendulum_noisy}
    \end{subfigure}
    \hfill
    \begin{subfigure}[b]{0.24\textwidth}
         \centering
         \includegraphics[width=\textwidth]{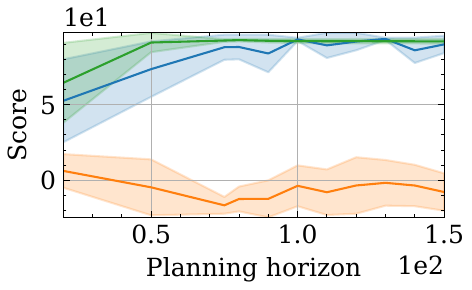}
         \caption{\scriptsize{CMC $(\alpha=0.002)$}}
         \label{fig:exp_depth_mountain_car_noisy}
    \end{subfigure}
    \hfill
    \begin{subfigure}[b]{0.24\textwidth}
         \centering
         \includegraphics[width=\textwidth]{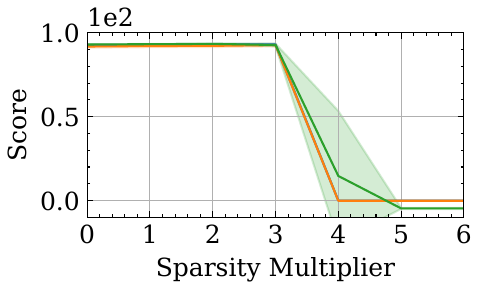}
         \caption{\scriptsize{CMC $(D=100)$}}
         \label{fig:exp_4_depth_100}
    \end{subfigure}
    \hfill
    \begin{subfigure}[b]{0.24\textwidth}
         \centering
         \includegraphics[width=\textwidth]{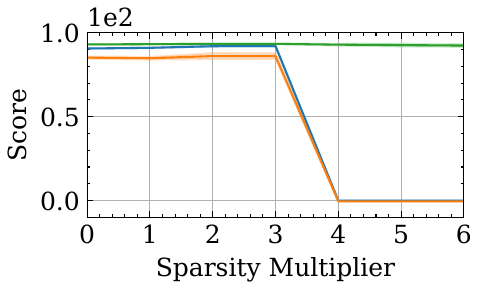}
         \caption{\scriptsize{CMC $(D=200)$}}
         \label{fig:exp_4_depth_200}
    \end{subfigure}
   % Row 3
    \begin{subfigure}[b]{0.24\textwidth}
         \centering
         \includegraphics[width=\textwidth]{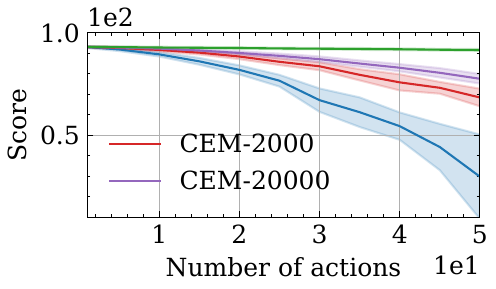}
         \caption{\scriptsize{CMC-HD $(D=100, \alpha=0)$}}
         \label{fig:exp_high_dim_scores}
    \end{subfigure}
    \hfill
    \begin{subfigure}[b]{0.24\textwidth}
         \centering
         \includegraphics[width=\textwidth]{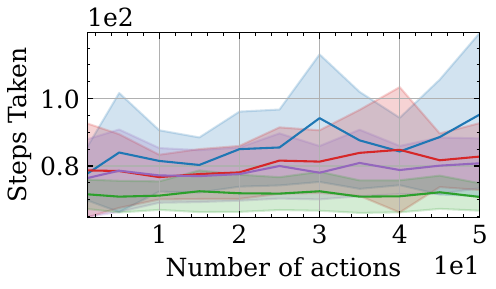}
         \caption{\scriptsize{CMC-HD $(D=100, \alpha=0)$}}
         \label{fig:exp_high_dim_steps}
    \end{subfigure}
    \hfill
    \begin{subfigure}[b]{0.24\textwidth}
         \centering
         \includegraphics[width=\textwidth]{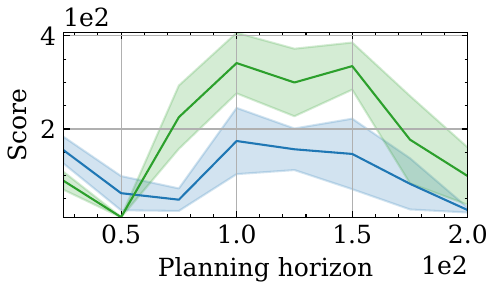}
         \caption{\scriptsize{CCP-Hybrid $(\alpha=0)$}}
         \label{fig:exp_hybrid_depth}
    \end{subfigure}
    \hfill
    \begin{subfigure}[b]{0.24\textwidth}
         \centering
         \includegraphics[width=\textwidth]{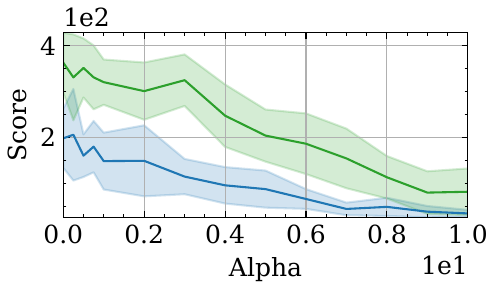}
         \caption{\scriptsize{CCP-Hybrid $(D=120)$}}
        \label{fig:exp_hybrid_noise}
    \end{subfigure}
  \\ 
  \caption{Environments used are Continuous Cartpole (CCP, $lr_\mu = 10$), Pendulum (P, $lr_\mu = 1$) and Continuous Mountain Car (CMC, $lr_\mu = 0.1$). \ref{fig:exp_noise_cartpole}, \ref{fig:exp_noise_pendulum}, \ref{fig:exp_noise_mountain_car}:
  While the performance of planners is similar in deterministic environments $(\alpha=0)$, \disprodSpace degrades more gracefully as compared to CEM and MPPI as the environments becomes more stochastic.
  %(Figures \ref{fig:exp_noise_cartpole}, \ref{fig:exp_noise_pendulum}, \ref{fig:exp_noise_mountain_car}). 
  \ref{fig:exp_depth_cartpole_noisy}, \ref{fig:exp_depth_pendulum_noisy}, \ref{fig:exp_depth_mountain_car_noisy}:
  In stochastic environments, using a large planning horizon can negatively impact the performance of CEM and MPPI while \disprodSpace gives better results.
  %(Figures \ref{fig:exp_depth_cartpole_noisy}, \ref{fig:exp_depth_pendulum_noisy}, \ref{fig:exp_depth_mountain_car_noisy}). 
  \ref{fig:exp_4_depth_100}, \ref{fig:exp_4_depth_200}:
  With a planning horizon of 100, all planners fail to reach the goal when the reward becomes very sparse, but \disprodSpace is able to recover by increasing the horizon to 200. 
  %(Figures \ref{fig:exp_4_depth_100}, \ref{fig:exp_4_depth_200}). 
  \ref{fig:exp_high_dim_scores}, \ref{fig:exp_high_dim_steps}:
  In CMC with large action space, CEM requires 100 times the number samples to achieve comparable results to \disprodSpace with 200 restarts. 
  %(Figures \ref{fig:exp_high_dim_scores}, \ref{fig:exp_high_dim_steps}). 
  \ref{fig:exp_hybrid_depth}, \ref{fig:exp_hybrid_noise}:
  \disprodSpace is able to achieve high rewards in hybrid settings as well.
  %(Figures \ref{fig:exp_hybrid_depth}, \ref{fig:exp_hybrid_noise})
  }
  \label{fig:exp_gym_planning_results}%
\end{figure*}

In the experiments we use 200 restarts, and for \disprodSpace $lr_v = \nicefrac{lr_\mu}{10}$. The values of $D$ and $lr_\mu$ for Gym environments are provided in \Cref{fig:exp_gym_planning_results}.
For TurtleBot we use $D=100$, $lr_\mu=10$. For Jackal and Heron, $D$ is modified to 30, 70 respectively, and we use 400 restarts.

\subsection{Baselines}
We compare \disprodSpace to CEM
and MPPI.
Both 
are shooting-based planners that maintain a sequence of action distributions to sample actions from.
From a given state, they use the sampled actions to generate multiple trajectories and compute the rewards accumulated at the end of each trajectory.
The two algorithms differ in
how they recompute the action distribution.
While CEM uses the actions 
from the
the top $n$ trajectories to form the new action distribution, MPPI 
weights the actions in a trajectory by the cumulative reward for that trajectory.
Hyperparameters for the planners were frozen after tuning them on deterministic versions of the environments.

CEM and MPPI can potentially benefit from the use of saving actions.
We have found that this is helpful in the basic environments but harms their performance 
in the car model. 
To ensure a fair comparison, we use the best setting for the baselines 
whereas \disprodSpace always saves actions.

\subsection{Evaluation in Basic Gym Environments}

\paragraph{Increasing Stochasticity.}
First, we explore the performance of the planners with the original deterministic environments and with added stochasticity.
To separate randomness in the environment from randomness in the experiment, we perform 8 repetitions over 6 runs in each environment, calculating means in each repetition and standard deviations of the means across repetitions.
Averages and standard deviations are shown in \Cref{fig:exp_noise_cartpole,fig:exp_noise_pendulum,fig:exp_noise_mountain_car}. We observe that the planners perform similarly in deterministic environments ($\alpha=0$) but 
with increasing amounts of stochasticity,
\disprodSpace degrades more gracefully and performs better than CEM or MPPI.

\paragraph{Increasing Planning Horizon.}
Intuitively the deeper the search, the more informative it is. But for sampling-based planners, deeper search can also increase variance when estimating action quality and hence harm performance. 
Results for experiments testing this aspect 
in noisy variants of basic environments 
are shown in \Cref{fig:exp_depth_cartpole_noisy,fig:exp_depth_pendulum_noisy,fig:exp_depth_mountain_car_noisy} and deterministic environments are included in the full paper.
We observe that while in the deterministic setting all the planners have similar performance, in some stochastic environments, the performance of CEM and MPPI can degrade with increasing search depth while \disprodSpace gives better results.
In these environments, beyond a required minimum depth, increasing the depth further does not help performance.
This changes, however, when we change reward sparsity.

\paragraph{Increasing Reward Sparsity.}
\label{sec:planning_with_sparse_rewards}
Robust planners should be able to work with both sparse and dense rewards.
Intuitively, if the reward is dense then even a short search horizon can yield good performance, but a deeper search is required for sparse rewards.
To test this, we evaluate the performance of the planners by varying the sparsity of the rewards in the Mountain Car environment. With the standard reward function, the agent gets a large positive reward when its position and velocity are larger than certain thresholds. We modify the reward function to use a smooth version of greater-than-equal-to function given by $\sigma(10\beta(x-\text{target}))$
where $\beta$ is a sparsity multiplier ($\beta=1$ in earlier experiments) and $\sigma(a)=1/(1+e^{-a})$.
The larger the value of $\beta$, the harder it is to get a reward from bad trajectories.
Results are shown in \Cref{fig:exp_4_depth_100,fig:exp_4_depth_200}.
\Cref{fig:exp_4_depth_100} shows that with the standard search depth of 100, all planners fail to reach the goal once the reward becomes very sparse.
\Cref{fig:exp_4_depth_200} shows that \disprodSpace can recover and perform well by increasing search depth to 200 but CEM and MPPI fail to do so.

\subsection{Evaluation in High Dimensional Action Space}
\label{sec: experiment_with_high_dim_action_space}
To explore 
high dimensional action spaces, 
we modify the Mountain Car environment by adding redundant action variables.
The dynamics still depend on a single relevant action variable but the reward function includes additional penalties for the redundant actions.
To obtain a high score, the agent must use a similar policy as before for the relevant action and keep the values of the redundant action variables as close to 0 as possible. Details of the model are in the full paper. We compare the performance of the planners against number of redundant actions, without changing any other hyperparameters.
Results are shown in \Cref{fig:exp_high_dim_scores}, where the MPPI results are omitted since the scores are too low and they distort the plot.
We observe that while CEM/MPPI perform poorly as the action space increases, the performance of \disprodSpace remains stable.
The performance of CEM improves if we increase the population size (number of samples) by a factor of 100 (from 200 to 20,000), but it still lags behind \disprodSpace. 
Additional results analyzing this scenario are in the full paper.
We note that despite the inferior reward, CEM is able to reach the goal location. However, as shown in \Cref{fig:exp_high_dim_steps}, it requires a lot more steps to reach the goal.
This experiment illustrates the potential advantage of planners that use 
the analytic model to identify what causes good outcomes 
as compared to estimating this effect through sampling.

\subsection{Evaluation with Hybrid State Space}
While we focus on experiments in continuous spaces, our planner is compatible with hybrid environments.  To illustrate this we modify the Cart Pole environment to include a binary variable which is set to 1 if the cart is to the right of a certain x-coordinate. The agent receives a reward of 3 when this binary variable is set to 1, otherwise it gets a reward of 1. 
\Cref{fig:exp_hybrid_depth,fig:exp_hybrid_noise} 
show the performance of \disprodSpace and CEM against planning horizon and noise level. 
We observe that a deeper search is required for this problem and that \disprodSpace can successfully obtain high reward.

\subsection{Evaluation with a Physics Simulator}
\label{sec:evaluation_with_a_physics_simulator}

We control a TurtleBot using 
\disprodSpace on 16 maps with varying degrees of difficulty. In these experiments, the analytic transition model does not take obstacles into account.
Incorporating obstacles in the transition yields similar or better results, but slows down the planner due to the increased size of the analytical computation graph. 
Instead, the 
reward function penalizes
the agent on collision with obstacles.
The obstacle patterns and detailed evaluation results are in the full paper.
Following \citeAY{wu2020towards}, we use success rate (SR) and success weighted by optimal path length (SL) for evaluation.
SR measures the success percentage across different maps, while SL is the ratio of actual path length to the euclidean distance from starting position to goal, averaged over cases where navigation was successful. Intuitively, the lower the SL value, the faster the agent reaches the goal.

Results are shown in \Cref{tab:turtlebotresults}, averaged over 5 runs for each map and averaged over all instances. We observe that all planners are able to control the TurtleBot but \disprodSpace performs better in both metrics.
We also evaluated the planners in a Gym environment where the planners' model matches the environment dynamics exactly. In this case, all planners perform similarly. 
Hence performance differences in TurtleBot are mainly due to better handling of the inaccurate model.

\subsection{Experiments with Robotic Systems}
\label{sec: exp_with_robotics_systems}
We use \disprodSpace with the same analytic model 
to control Jackal and Heron. Real-time control requires high frequency control commands to avoid significant drift. Therefore, one has to
find a good balance between planning horizon $(D)$ and the maximum change of linear velocity $\overline{\Delta{v}}$.
Experiments with low $\overline{\Delta{v}}$ require a large $D$ and are slow, while experiments with high $\overline{\Delta{v}}$ suffer from drift, as the hardware cannot stop or accelerate instantaneously.
We specified these parameters through initial exploration with the systems. Other parameters are the same as in the TurtleBot simulation.

\begin{table}[tb]
  \centering
  \scriptsize
  \begin{tabular}{rrrr}
    \toprule
    {Environments} & {Method} & {Success Rate~(SR)$\uparrow^{100}_{0}$} & {Success Length~(SL)$\downarrow$}\\
    \midrule
    OpenAI Gym    & CEM       &     $\boldsymbol{100.00}$   & $\boldsymbol{1.36}$\\
                  & MPPI      &     $97.30$                 & $1.42$\\
                  & DiSProD   &     $\boldsymbol{100.00}$   & $1.44$\\
                  \midrule
    TurtleBot     & CEM       &     $85.88$                 & $1.56$\\
                  & MPPI      &     $85.88$                 & $1.62$\\
                  & DiSProD   &     $\boldsymbol{95.29}$    & $\boldsymbol{1.54}$\\
                  \bottomrule
  \end{tabular}
  \caption{Aggregated SR and SL for all maps when using the Dubins car model to plan in our Gym simulator and TurtleBot.}\label{tab:turtlebotresults}%
\end{table}

\begin{figure}[tbp]%
    \centering%
     \subfloat{%
        \includegraphics[width=0.7\linewidth]{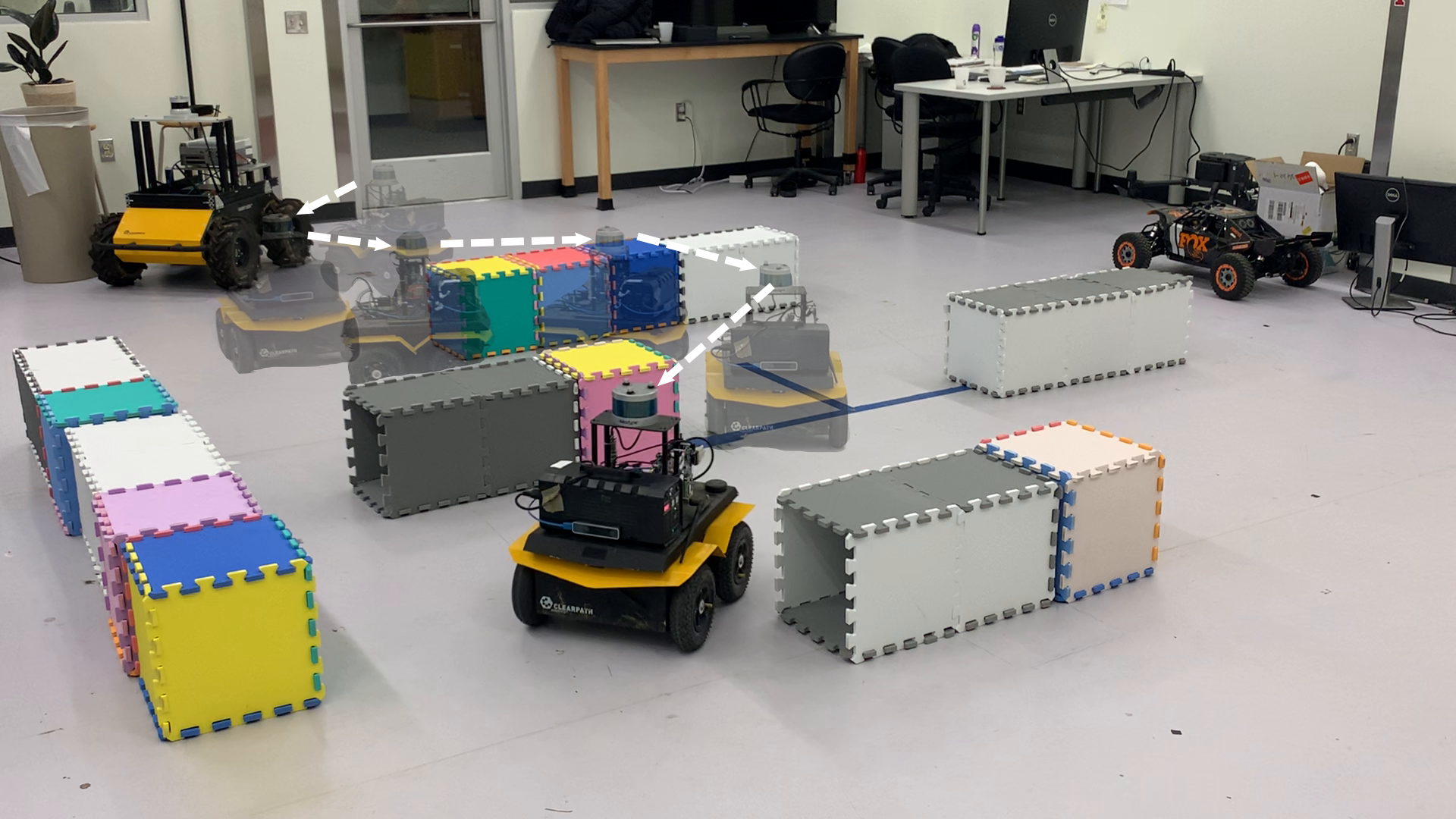}%
    }\\
    \subfloat{%
        \includegraphics[width=0.7\linewidth]{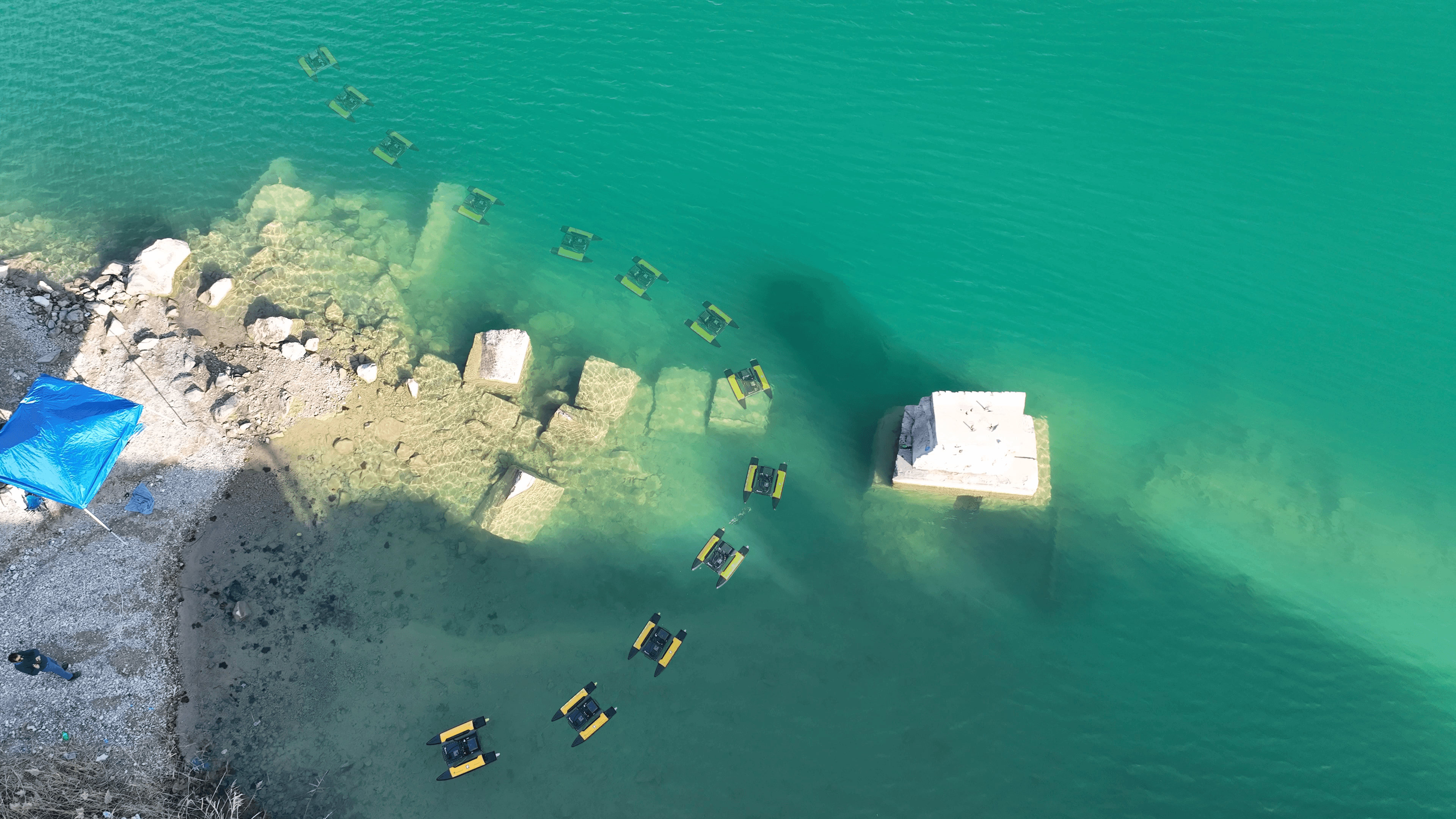} %
    }
   
    \caption{DiSProD controls an Unmanned Ground Vehicle and an Unmanned Surface Vessel to navigate around obstacles. The opacity indicates the robot's poses at different times.}\label{fig:exp_hardware_results}%
\end{figure}

For experiments with Jackal, \disprodSpace is used exactly as above, i.e., sending actions to the robot.
For the surface vessel, however, we use our planner in another way because our Heron has a motor issue in one of its thrusters.
Specifically, \disprodSpace optimizes the action sequence exactly as before.
Instead of sending the actions, it computes intermediate states expected to be reached with the policy (which are available in the computation graph), and sends these as ``waypoints" to a PID controller.
We found that, since our planner works at a fine time granularity, we can send every 5th state to a PID controller and achieve smooth control.
\Cref{fig:exp_hardware_results} visualizes the trajectories generated by \disprodSpace when controlling Jackal and Heron. 
Some videos from these experiments can be seen
at \url{https://pecey.github.io/DiSProD}.

\begin{figure}[tb]%
    \centering
    % \subfloat{%
    %    \includegraphics[width=0.7\linewidth]{figures/images/visualizing_rollouts/legend.pdf}
    % }\\ 
    % \subfloat{%
    %     \includegraphics[width=0.5\linewidth]{figures/images/visualizing_rollouts/rollouts_1.pdf}
    % }
    % \subfloat{%
    %     \includegraphics[width=0.5\linewidth]{figures/images/visualizing_rollouts/rollouts_2.pdf}
    % }
    \subfloat{%
        \includegraphics[width=0.9\linewidth]{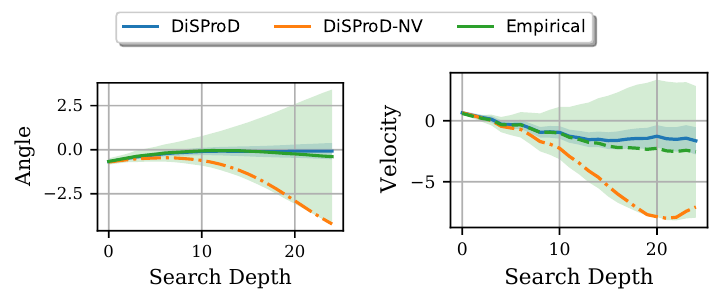}
    }
    \caption{Comparing empirical state distributions in Pendulum ($\alpha =1)$ with approximations computed by DiSProD and DiSProD-NV. %Shown in color: empirical  (green), \disprodSpace (blue), \disprod-NV (orange). 
    }\label{fig:rollouts_stochastic_simple_env_fixed_alpha}%
\end{figure}

\subsection{Ablation Study and Runtime Comparison}

We first evaluate
the contribution of the different variance terms $(\hat{v}_{{\bm{s}}_t}, \hat{v}_{{\bm{a}}_t}$ and $\hat{v}_{\bm{\epsilon}_t})$ in \Cref{eq:expectation_vector_form,eq:variance_vector_form}
to the performance of the algorithm.
We use the term \textit{complete mode} when the planner uses the variance terms (\disprod) and \textit{no-variance mode} when it zeroes them out (\disprod-NV).
We look into the state distributions produced by \disprodSpace and \disprod-NV, and compare them against empirical state distributions.  
We fix the start state and a sequence of action distributions, and compute the next-state distribution for a fixed depth. 
For the empirical state distribution, 
we sample actions from the same fixed action distribution and use the dynamics model to compute the next-state.
The trajectory distributions for Pendulum are visualized in \Cref{fig:rollouts_stochastic_simple_env_fixed_alpha}.
We observe that \disprodSpace gives us a better approximation than \disprod-NV -- the mean is more accurate and while the variance is underestimated, it has a reasonable shape.
Additional experiments
in the full paper show additional visualization of trajectory distributions, as well as showing that both action and state variance 
contribute to the improved planning performance.

We next consider run-time comparing \disprodSpace to the baselines. 
Evaluating on the basic Gym environments (details in the full paper),
\disprod-NV has roughly the same run time as CEM and MPPI and \disprodSpace is up to 7 times slower. 
This gap is expected to increase with 
more state variables
due to the inclusion of partial derivatives in the computation graph.  
\section{Conclusion}

The paper presents \disprod, a novel approach for planning in continuous stochastic environments.
The method is general and overcomes limitations of prior work by using an abstracted representation, using a higher order Taylor approximation, and showing how optimization via gradients can be done over propagation of distributions.
Experiments show success across multiple problems, improved handling of stochastic environments, and decreased sensitivity to search depth, reward sparsity, and large action spaces.
\disprodSpace is also shown to be compatible with control in real robotic systems.
At present, the key limitations of \disprodSpace are its computational complexity arising from incorporating the first and second order partials in the computation graph, over which we perform gradient search,
its approximation quality requiring non-zero partial derivatives,
and the need for a known model which may be resolved using model-based RL.
These are left as important questions for future work.

\clearpage
\newpage

\section*{Acknowledgements}
We are grateful to members of the VAIL Lab at Indiana University, especially Durgakant Pushp and Lantao Liu, for
helping with the robotic experiments. This work was partly supported by NSF under grants 2002393, 2006886, 2047169 and 2246261. Some of the
experiments in this paper were run on the Big Red computing system at Indiana University, supported in part by Lilly
Endowment, Inc., through its support for the Indiana University Pervasive Technology Institute.

\bibliographystyle{named}
\bibliography{refs}

\clearpage
\newpage

\appendix
\section*{Technical Appendix}
\section{Example: Computation Graphs with Known Transition Models}

For a concrete example, let us consider the dynamics of the Pendulum from the OpenAI Gym. Let $S_t = [\theta_t, \Dot{\theta}_t]^\top$ be the state vector, $A_t = [a_t]^\top$ be the action vector and $\epsilon_t = [e_t]^\top$ be the noise vector. The transition function and the reward function are defined as follows: 
\begin{align}
    T(S,A, \epsilon) &= [\theta_{t+1}, \Dot{\theta}_{t+1}]^\top, \\ 
    \text{where }  \Dot{\theta}_{t+1} &= \Dot{\theta}_{t} + \left(-c_1 \sin(\theta_t + \pi) +  c_2 a_t\right) \Delta_t \nonumber \\
    \theta_{t+1} &= \theta_t + (\Dot{\theta}_{t+1} + e_t) \Delta_t \nonumber \\
    \mathcal{R}(S,A) &= - \theta_t^2 - 0.1 \times \Dot{\theta_t}^2 - 0.001 \times a_t^2 \\
\end{align}
The original reward function uses a normalized $\theta_t$, but we ignore it for this example, in order to keep the computations simple.
Here, $c_1 = \frac{3g}{2l}$ and $c_2 = \frac{3}{ml^2}$, are constants.
Then the expectation and variance of the transition function can be written as follows:
\begin{align}
% FOP wrt state
   J^{T}_{S_t} 
   &= \frac{\partial S_{t+1}}{\partial S_{t}}
    =     \begin{bmatrix}
         \frac{\partial \theta_{t+1}} {\partial \theta_{t}} &
         \frac{\partial \theta_{t+1}} {\partial \Dot{\theta}_{t}} \\
         \frac{\partial \Dot{\theta}_{t+1}} {\partial \theta_{t}} & 
         \frac{\partial \Dot{\theta}_{t+1}} {\partial \Dot{\theta}_{t}}
    \end{bmatrix} \nonumber \\
    &= \begin{bmatrix}
         1 - c_1 \cos(\theta_t + \pi) {\Delta^2_t} &
         \Delta_t \\
         -c_1 \cos(\theta_t + \pi) {\Delta_t} & 
         1
    \end{bmatrix}\\
% FOP wrt action
    J^{T}_{A_t} 
    &= \frac{\partial S_{t+1}}{\partial A_{t}}
    = \begin{bmatrix}
         \frac{\partial \theta_{t+1}} {\partial a_{t}}\\
         \frac{\partial \Dot{\theta}_{t+1}} {\partial a_{t}} 
    \end{bmatrix}
    = \begin{bmatrix}
         c_2 \Delta^2_t\\
         c_2 \Delta_t
    \end{bmatrix}\\
% FOP wrt error
    J^{T}_{\epsilon_t} 
    &= \frac{\partial S_{t+1}}{\partial \epsilon_{t}}
    = \begin{bmatrix}
         \frac{\partial \theta_{t+1}} {\partial e_{t}} \\
         \frac{\partial \Dot{\theta}_{t+1}} {\partial e_{t}} 
    \end{bmatrix}
    = \begin{bmatrix}
         \Delta_t \\
         0
    \end{bmatrix}\\
% SOP wrt state
    \tilde{H}^{T}_{S_t} 
    &= \frac{\partial^2 S_{t+1}}{\partial S_{t}^2} 
    = \begin{bmatrix}
         \frac{\partial^2 \theta_{t+1}} {\partial \theta_{t}^2} &
         \frac{\partial^2 \theta_{t+1}} {\partial \Dot{\theta}_{t}^2} \\
         \frac{\partial^2 \Dot{\theta}_{t+1}} {\partial \theta_{t}^2} & 
         \frac{\partial^2 \Dot{\theta}_{t+1}} {\partial \Dot{\theta}_{t}^2}
        \end{bmatrix} \nonumber \\
    &=  \begin{bmatrix}
         c_1 \sin(\theta_t + \pi) \Delta^2_t &
         0 \\
         c_1 \sin(\theta_t + \pi) \Delta_t & 
         0
        \end{bmatrix}\\
% SOP wrt action
    \tilde{H}^{T}_{A_t} 
    &= \frac{\partial^2 S_{t+1}}{\partial A_{t}^2}
    = \begin{bmatrix}
             \frac{\partial^2 \theta_{t+1}} {\partial a_{t}^2} \\
             \frac{\partial^2 \Dot{\theta}_{t+1}} {\partial a_{t}^2} 
        \end{bmatrix}
    =  \begin{bmatrix}
             0 \\
             0
        \end{bmatrix}\\
% SOP wrt error
    \tilde{H}^{T}_{\epsilon_t} 
    &= \frac{\partial^2 S_{t+1}}{\partial \epsilon_{t}^2}
    = \begin{bmatrix}
             \frac{\partial^2 \theta_{t+1}} {\partial e_{t}^2} \\
             \frac{\partial^2 \Dot{\theta}_{t+1}} {\partial e_{t}^2} 
            \end{bmatrix}
    =   \begin{bmatrix}
             0 \\
             0 
        \end{bmatrix}\\
% Variance vectors
\hat{\bm{v}}_{S_t} 
    &= \begin{bmatrix}
         \hat{v}_{\theta, t} \\
         \hat{v}_{\Dot{\theta}, t}
        \end{bmatrix} \qquad
\hat{\bm{v}}_{A_t} 
= \begin{bmatrix}
         \hat{v}_{a, t}
        \end{bmatrix} \qquad
\hat{\bm{v}}_{\epsilon_t} 
=  \begin{bmatrix}
         \hat{v}_{e, t}
        \end{bmatrix} 
\end{align}
\begin{align}
    \mathbb{E}[T(S_t,A_t, \epsilon)] &\approx T(\mu_{S_t}, \mu_{A_t}, \mu_\epsilon) \nonumber \\
    &+ 0.5 \left( 
        % WRT state variables
         \tilde{H}^T_{S_t}
         \hat{\bm{v}}_{S_t}
        +
        % WRT action variables
         \tilde{H}^T_{A_t}
        \hat{\bm{v}}_{A_t}
        +
        % WRT noise variables
        \tilde{H}^T_{\epsilon_t} 
       \hat{\bm{v}}_{\epsilon_t}
    \right)  \\
    % Variance
    \mathbb{V}[T(S_t,A_t, \epsilon)] &\approx
    % WRT state variables
    (J^T_{S_t} \odot J^T_{S_t})
   \hat{\bm{v}}_{S_t}
        +
    % WRT action variables
    (J^T_{A_t} \odot J^T_{A_t})
    \hat{\bm{v}}_{A_t}  \nonumber \\
        &+ 
    % WRT noise variables
   (J^T_{\epsilon_t} \odot J^T_{\epsilon_t})
    \hat{\bm{v}}_{\epsilon_t}
\end{align}

Similarly, the expressions for the reward function are:
\begin{align}
    % SOP wrt state
    \tilde{H}^{R}_{S_t} 
    &= \begin{bmatrix}
         \frac{\partial^2 \mathcal{R}} {\partial \theta_{t}^2} &
         \frac{\partial^2 \mathcal{R}} {\partial \Dot{\theta}_{t}^2} 
        \end{bmatrix} 
    =  \begin{bmatrix}
         -2 &
         -0.2 \\
        \end{bmatrix}\\
% SOP wrt action
    \tilde{H}^{R}_{A_t} 
    &= \begin{bmatrix}
             \frac{\partial^2 \mathcal{R}} {\partial a_{t}^2} 
        \end{bmatrix}
    =  \begin{bmatrix}
            -0.002
        \end{bmatrix}\\
% Reward expectation
    \mathbb{E}[\mathcal{R}(S_t,A_t)] &\approx \mathcal{R}(\mu_{S_t}, \mu_{A_t}) \nonumber \\
    &+ 0.5 \left( 
        % WRT state variables
         \tilde{H}^R_{S_t}
         \hat{\bm{v}}_{S_t}
        +
        % WRT action variables
         \tilde{H}^R_{A_t}
        \hat{\bm{v}}_{A_t}
    \right) 
\end{align}
This illustrates that we can have analytic gradients for the model as mentioned above. 
Note that, we do not model the variance of the reward function as mentioned in \Cref{sec: analytic_graph}. For the sake of simplicity, we have considered just a single noise variable in this model. However, the same approach works with multiple noise variables. 

% \begin{wrapfigure}{r}{0.5\textwidth}
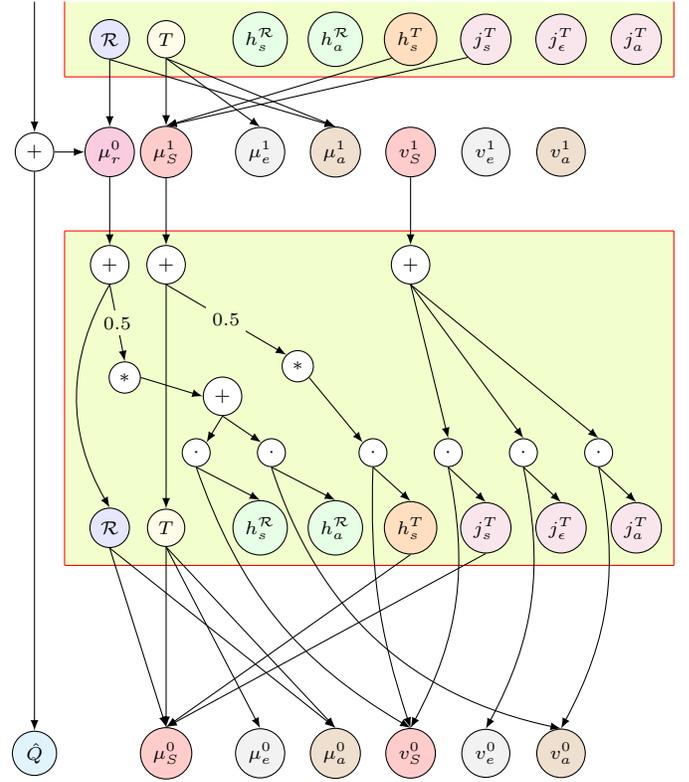
\begin{figure}[t]%
  \centering%
  % \resizebox{\linewidth}{!}{%
  \begin{tikzpicture}[]\scriptsize%
    % Rect
    \draw [draw=lime!20,fill=lime,opacity=0.2] (0.4,1.5) rectangle (8.5,5.95);
    %\draw [draw=lime!20,fill=lime,opacity=0.2] (0.4,6.5) rectangle (8.5,9);
    \draw [draw=lime!20,fill=lime,opacity=0.2] (0.4,8.0) rectangle (8.5,9);
    % Edges of bottom rect
    \draw [-,draw=red] (0.4,1.5) to (8.5,1.5);
    \draw [-,draw=red] (0.4,1.5) to (0.4,5.95);
    \draw [-,draw=red] (0.4,5.95) to (8.5,5.95);
    \draw [-,draw=red] (8.5,5.95) to (8.5,1.5);

    % Edges of top rect
    %\draw [-,draw=red] (0.4,6.5) to (0.4,9);
    %\draw [-,draw=red] (0.4,6.5) to (8.5,6.5);
    %\draw [-,draw=red] (8.5,6.5) to (8.5,9);
    \draw [-,draw=red] (0.4,8.0) to (0.4,9);
    \draw [-,draw=red] (0.4,8.0) to (8.5,8.0);
    \draw [-,draw=red] (8.5,8.0) to (8.5,9);
    
    % Nodes
    \node[circle,draw=black,fill=cyan!10] at (0,  -1) (0_Q)     {$\hat{Q}$};%
    %\node[circle,draw=black] at (1,  -1) (0_R)     {$R$};%
    \node[circle,draw=black,fill=red!20] at (1.75,  -1) (0_S_mu)  {${\mu}_S^0$};%
    \node[circle,draw=black,fill=gray!10] at (3,  -1) (0_e_mu)  {${\mu}_e^{0}$};%
    \node[circle,draw=black,fill=brown!25] at (4,  -1) (0_a_mu)  {${\mu}_a^{0}$};%
    \node[circle,draw=black,fill=red!20] at (5,  -1) (0_S_nu)  {${v}_S^0$};%
    \node[circle,draw=black,fill=gray!10] at (6,  -1) (0_e_nu)  {${v}_e^{0}$};%
    \node[circle,draw=black,fill=brown!25] at (7,  -1) (0_a_nu)  {${v}_a^{0}$};%
    %
    %\node[circle,draw=black]                at (0,  2) (2_+)  {$+$};%
    \node[circle,draw=black,fill=blue!10]   at (1,  2) (2_r)  {$\mathcal{R}$};%
    \node[circle,draw=black,fill=yellow!10] at (1.75,  2) (2_T)  {$T$};%
    \node[circle,draw=black,fill=green!10]  at (3,  2) (2_hr_1) {$h^\mathcal{R}_{s}$};%
    \node[circle,draw=black,fill=green!10]  at (4,  2) (2_hr_2) {$h^\mathcal{R}_{a}$};%
    \node[circle,draw=black,fill=orange!25] at (5,  2) (2_h_1) {$h^T_{s}$};%
    \node[circle,draw=black,fill=purple!10]  at (6,  2) (2_j_1) {$j^T_{s}$};%
    \node[circle,draw=black,fill=purple!10]  at (7,  2) (2_j_3) {$j^T_{\epsilon}$};%
    \node[circle,draw=black,fill=purple!10]  at (8,  2) (2_j_2) {$j^T_{a}$};%

    \node[circle,draw=black,fill=white]  at (2.15,  3) (3_r_1) {$\cdot$};%
    \node[circle,draw=black,fill=white]  at (3.15,  3) (3_r_2) {$\cdot$};%
    \node[circle,draw=black,fill=white]  at (4.5,  3) (3_1) {$\cdot$};%
    \node[circle,draw=black,fill=white]  at (5.5,  3) (3_2) {$\cdot$};%
    \node[circle,draw=black,fill=white]  at (6.5,  3) (3_3) {$\cdot$};%
    \node[circle,draw=black,fill=white]  at (7.5,  3) (3_4) {$\cdot$};%

     \node[circle,draw=black,fill=white]  at (1.2,  4.0) (3_0_*) {$*$};%
    \node[circle,draw=black,fill=white]  at (3.5,  4.15) (3_1_*) {$*$};%
    \node[circle,draw=black,fill=white]  at (2.5,  3.75) (3_+) {$+$};%

    \node[circle,draw=black,fill=white] at (1,  5.5) (4_+_0)  {$+$};%
    \node[circle,draw=black,fill=white] at (1.75,  5.5) (4_+_1)  {$+$};%
    \node[circle,draw=black,fill=white] at (5,  5.5) (4_+_2)  {$+$};%

    \node[circle,draw=black]                at (0,  7) (r_mu_0_+)  {$+$};%
    \node[circle,draw=black,fill=magenta!25] at (1,  7) (4_r_mu)  {${\mu}^{0}_r$};%
    \node[circle,draw=black,fill=red!20] at (1.75,  7) (4_s_mu)  {$\mu^{1}_S$};%
    \node[circle,draw=black,fill=gray!10] at (3,  7) (4_e_mu)  {${\mu}_e^{1}$};%
    \node[circle,draw=black,fill=brown!25] at (4,  7) (4_a_mu)  {${\mu}_a^{1}$};%
    \node[circle,draw=black,fill=red!20] at (5,  7) (4_s_v)  {${v}_S^{1}$};%
    \node[circle,draw=black,fill=gray!10] at (6,  7) (4_e_nu)  {${v}_e^{1}$};%
    \node[circle,draw=black,fill=brown!25] at (7,  7) (4_a_nu)  {${v}_a^{1}$};%
    %
    % \node[circle,draw=black]                at (0,  8.5) (6_+)  {$+$};%
    \node[circle,draw=black,fill=blue!10]   at (1,  8.5) (6_r)  {$\mathcal{R}$};%
    \node[circle,draw=black,fill=yellow!10] at (1.75,  8.5) (6_T)  {$T$};%
    \node[circle,draw=black,fill=green!10]  at (3,  8.5) (6_hr_1) {$h^\mathcal{R}_{s}$};%
    \node[circle,draw=black,fill=green!10]  at (4,  8.5) (6_hr_2) {$h^\mathcal{R}_{a}$};%
    \node[circle,draw=black,fill=orange!25] at (5,  8.5) (6_h_1) {$h^T_{s}$};%
    \node[circle,draw=black,fill=purple!10]  at (6,  8.5) (6_j_1) {$j^T_{s}$};%
    \node[circle,draw=black,fill=purple!10]  at (7,  8.5) (6_j_3) {$j^T_{\epsilon}$};%
    \node[circle,draw=black,fill=purple!10]  at (8,  8.5) (6_j_2) {$j^T_{a}$};%

    %  Edges
    %\draw [-latex] (2_+)         to node {} (2_r);
    \draw [-latex] (2_r.south)   to node {} (0_S_mu.north);
    \draw [-latex] (2_r.south)   to node {} (0_a_mu.north);
    % \draw [-latex] (2_r.south)   to node {} (0_S_nu.north);
    % \draw [-latex] (2_r.south)   to node {} (0_a_nu.north);
    \draw [-latex] (2_T.south)   to node {} (0_S_mu.north);
    \draw [-latex] (2_T.south)   to node {} (0_e_mu.north);
    \draw [-latex] (2_T.south)   to node {} (0_a_mu.north);
    \draw [-latex] (2_h_1.south) to node {} (0_S_mu.north);
    \draw [-latex] (2_j_1.south) to node {} (0_S_mu.north);
    % cdot to partials
    \draw [-latex] (3_r_1.south) to node {} (2_hr_1.north);
    \draw [-latex] (3_r_2.south) to node {} (2_hr_2.north);
    \draw [-latex] (3_1.south) to node {} (2_h_1.north);
    \draw [-latex] (3_2.south) to node {} (2_j_1.north);
    \draw [-latex] (3_3.south) to node {} (2_j_3.north);
    \draw [-latex] (3_4.south) to node {} (2_j_2.north);
    % cdot to variance
    \draw [-latex] (3_r_1.south) to [bend right=20] node {} (0_S_nu.north);
    \draw [-latex] (3_r_2.south) to [bend right=30] node {} (0_a_nu.north);
    % \draw [-] (4,1.5) to node {} (5,1);
    % \draw [-latex] (5,1) to  node {} (0_a_nu.north west);
    \draw [-latex] (3_1.south) to [bend right=10] node {} (0_S_nu.north);
    \draw [-latex] (3_2.south) to [bend left=20] node {} (0_S_nu.north);
    \draw [-latex] (3_3.south) to [bend left=20] node {} (0_e_nu.north);
    \draw [-latex] (3_4.south) to [bend left=20] node {} (0_a_nu.north);
    \draw [-latex] (4_+_1.south) to node {} (2_T.north);
    \draw [-latex] (4_+_1.south) to node [fill=lime!20] {$0.5$}  (3_1_*.north west);
    \draw [-latex] (3_1_*.south east) to node {}  (3_1.north west);
    %\draw [-latex] (3_*.west) to node {}  (3_+.east);
    \draw [-latex] (4_+_0.south) to [bend right=30] node  {}  (2_r.north);
    \draw [-latex] (4_+_0.south) to node [fill=lime!20] {$0.5$}  (3_0_*.north);
    \draw [-latex] (3_0_*.east) to node {}  (3_+.west);
    \draw [-latex] (3_+.south) to node {}  (3_r_1.north east);
    \draw [-latex] (3_+.south) to node {}  (3_r_2.north west);
    % \draw [-latex] (4_+_1.south) to node [fill=white] {$0.5*$}  (3_1.north);
    \draw [-latex] (4_+_2.south) to node {} (3_2.north);
    \draw [-latex] (4_+_2.south) to node {} (3_3.north);
    \draw [-latex] (4_+_2.south) to node {} (3_4.north);

    \draw[-latex] (4_r_mu.south) to node{} (4_+_0.north);
    \draw[-latex] (4_s_mu.south) to node{} (4_+_1.north);
    \draw[-latex] (4_s_v.south) to node{} (4_+_2.north);
    
    %\draw [-latex] (6_+)    to node {} (6_r);
    % \draw [-latex] (6_r.south)    to node {} (4_+_1.north);
    \draw [-latex] (6_r.south)    to node {} (4_a_mu.north);
    \draw [-latex] (6_r.south)    to node {} (4_r_mu.north);
    %\draw [-latex] (6_r.south)    to node {} (4_+_2.north);
    %\draw [-latex] (6_r.south)    to node {} (4_a_nu.north);
    \draw [-latex] (6_T.south)    to node {} (4_s_mu.north);
    \draw [-latex] (6_T.south)    to node {} (4_e_mu.north);
    \draw [-latex] (6_T.south)    to node {} (4_a_mu.north);
    \draw [-latex] (6_h_1.south west)  to node {} (4_s_mu.north);
    \draw [-latex] (6_j_1.south west)  to node {} (4_s_mu.north);
     %\draw [-latex] (6_hr_1.south)  to node {} (4_+_1.north);
    % Q fn
     \draw [-latex] (r_mu_0_+.south)  to node {} (0_Q.north);
      \draw [-latex] (0, 9)  to node {} (r_mu_0_+.north);
     \draw [-latex] (r_mu_0_+.east)  to node {} (4_r_mu.west);
  \end{tikzpicture}%
  % }%
  \caption{Computational diagram of the proposed planner for Pendulum 
  showing the distribution block for depth 1 and a portion of the block for depth 2, 
  as well as  showing how the reward is accumulated in $\hat{Q}$.
  Node $T$ and $\mathcal{R}$ represents the transition and reward function respectively. 
  $\mu^0_S = S_0$ and $v^0_S = 0$ capture the initial state distribution. 
  $\mu^{i}_a$ and $v^{i}_a$ capture the action distribution at depth $i$. 
  Similarly $\mu^{j}_e = 0$ and $v^{j}_e = 1$ capture the error distribution at depth $j$. 
  Nodes $h^T_s$ and $j^T_s, j^T_{\epsilon}, j^T_a$ help in computing the next state mean and variance and node $h^\mathcal{R}_s$ and $h^\mathcal{R}_a$ help in computing the mean reward. 
  Note that $h^T_a, h^T_\epsilon$ are both zero and hence they are omitted from the graph.
  The nodes $h^\mathcal{R}_s$, $h^\mathcal{R}_a$, $j^T_{\epsilon}$ and $j^T_a$ have a constant value
  and therefore they do not have incoming edges. 
  }\label{fig:diagram}\vspace{-10pt}%
\end{figure}
% \end{wrapfigure}

The computation graph 
for this example is shown in Figure \ref{fig:diagram}. To make the graph simple, certain portions have been abstracted away. More specifically, $T$ signifies the transition function and $\mathcal{R}$ specifies the reward function. 
For this example, nodes $h^T_s, j^T_s, j^T_a$ and $j^T_{\epsilon}$ are used to compute the expectation and variance of next state variables and are defined as follows: $h^T_s = \Tilde{H}^T_{S_t}$;
                            $j^T_s = J^T_{S_t} \odot J^T_{S_t}$;
                            $j^T_a = J^T_{A_t} \odot J^T_{A_t}$; 
                          $j^T_{\epsilon} = J^T_{\epsilon_t} \odot J^T_{\epsilon_t}$.
Similarly, nodes $h^{\mathcal{R}}_s$ and $h^{\mathcal{R}}_a$ are used to compute the expected reward and are defined as: $h^{\mathcal{R}}_s = \Tilde{H}^\mathcal{R}_{S_t}$; $h^{\mathcal{R}}_a = \Tilde{H}^\mathcal{R}_{A_t}$.

% !TEX root =  main.tex

\section{Contribution of Sub-Expressions in the Taylor's Expansion}
\label{sec:analysis_of_ablation_modes}

As described in Section \ref{sec: analytic_graph}, the vector form of the equations are as follows:

\begin{align}
% Expectation
    \hat{\mu}_{\bm{s}_{t+1}} &\approx T(\hat{\bm{z}_t}) + 0.5 \biggl[ \tilde{H}_{\bm{s}_t} \hat{v}_{\bm{s}_t} + \tilde{H}_{\bm{a}_t} \hat{v}_{\bm{a}_t} + \tilde{H}_{\bm{\epsilon}_t} \hat{v}_{\bm{\epsilon}_t} \biggr]\\
% Variance
\hat{v}_{\bm{s}_{t+1}}  &\approx (J_{\bm{s}_t} \odot J_{\bm{s}_t}) \hat{v}_{\bm{s}_t} + (J_{\bm{a}_t} \odot J_{\bm{a}_t}) \hat{v}_{\bm{a}_t} \\
&+ (J_{\bm{\epsilon}_t} \odot J_{\bm{\epsilon}_t}) \hat{v}_{\bm{\epsilon}_t} \nonumber
\end{align}
where $\hat{\bm{z}_t} = (\hat{\mu}_{\bm{s}_{t}}, {\mu}_{\bm{a}_{t}}, {\mu}_{\bm{\epsilon}_{t}})$

The equations depend on the distribution of state, action and noise variables. To analyze the impact each set of variables has on the outcome, we consider three cases - no-variance, state-variance and complete.

% No variance Mode
\mypar{No variance} In this setup, all the variance terms$(\hat{v}_{\bm{s}}, \hat{v}_{\bm{a}}, \hat{v}_{\bm{\epsilon}})$ are zeroed out. So the equations become
\begin{align}
    % Expectation
    \hat{\mu}_{\bm{s}_{t+1}}  &\approx T_j(\hat{\bm{z}}) \\
    % Variance
    \hat{v}_{\bm{s}_{t+1}}  &\approx 0
\end{align}

% State variance mode
\mypar{State variance} Now, we just zero out action variance $(\hat{v}_{\bm{a}})$, which means we effectively optimize a deterministic policy.

\begin{align}
% Expectation
    \hat{\mu}_{\bm{s}_{t+1}} &\approx T(\hat{\bm{z}_t}) + 0.5 \biggl[ \tilde{H}_{\bm{s}_t} \hat{v}_{\bm{s}_t} + \tilde{H}_{\bm{\epsilon}_t} \hat{v}_{\bm{\epsilon}_t} \biggr]\\
% Variance
\hat{v}_{\bm{s}_{t+1}}  &\approx (J_{\bm{s}_t} \odot J_{\bm{s}_t}) \hat{v}_{\bm{s}_t} + (J_{\bm{\epsilon}_t} \odot J_{\bm{\epsilon}_t}) \hat{v}_{\bm{\epsilon}_t}
\end{align}

Note that at $t=0$, as the planner knows the state exactly, therefore $\hat{v}_{\bm{s}_0} = 0$. We now consider the possible values the remaining partials might take and analyze the impact. 

\begin{enumerate}
    \item $J_{\bm{\epsilon}_t} = 0$ and $\tilde{H}_{\bm{\epsilon}_t} = 0$. Then,
    \begin{align}
        % Case 1: Expectation
        \hat{\mu}_{\bm{s}_1} \approx T(\hat{\bm{z}_0}) 
        \quad \text{and}\quad
        % Case 1: Variance
        \hat{v}_{\bm{s}_1} \approx 0
    \end{align}
    As $\hat{v}_{\bm{s}_t}$ always remains zero, the result is similar to the no-variance mode.
    
    \item $J_{\bm{\epsilon}_t} \ne 0$ but $\tilde{H}_{\bm{\epsilon}_t} = 0$. Then,
    \begin{align}
        % Case 2: Expectation
        \hat{\mu}_{\bm{s}_1}  \approx T(\hat{\bm{z}_0})
         \quad \text{and}\quad
        % Case 2: Variance
        \hat{v}_{\bm{s}_1} \approx (J_{\bm{\epsilon}_t} \odot J_{\bm{\epsilon}_t}) \hat{v}_{\bm{\epsilon}_t}
    \end{align}
    As $\hat{v}_{\bm{s}_1} \ne 0$, it influences $\hat{\mu}_{\bm{s}_2}$ and  $\hat{v}_{\bm{s}_2}$ and future timesteps.
    
    \item $J_{\bm{\epsilon}_t} = 0$ but $\tilde{H}_{\bm{\epsilon}_t} \ne 0$. This can happen as we evaluate the partials at the mean of the variables, and we assume zero-mean noise ($\hat{\mu}_{\bm{\epsilon}_t} = 0$). So while $J_{\bm{\epsilon}_t}$ exists, it can evaluate to zero. In such a scenario, we have,
    \begin{align}
        % Case 2: Expectation
        \hat{\mu}_{\bm{s}_1}  \approx T(\hat{\bm{z}_0}) + \tilde{H}_{\bm{\epsilon}_t} \hat{v}_{\bm{\epsilon}_t}
         \quad \text{and}\quad
        % Case 2: Variance
        \hat{v}_{\bm{s}_1} = 0
    \end{align}
    As $\hat{v}_{\bm{s}_{t+1}}  \approx (J_{\bm{s}_t} \odot J_{\bm{s}_t}) \hat{v}_{\bm{s}_t}$ and $\hat{v}_{\bm{s}_0} = 0$, the next-state variance is always zero. We can generalize and write 
     \begin{align}
        % Case 2: Expectation
        \hat{\mu}_{\bm{s}_{t+1}}  \approx T(\hat{\bm{z}_t}) + \tilde{H}_{\bm{\epsilon}_t} \hat{v}_{\bm{\epsilon}_t}
         \quad \text{and}\quad
        % Case 2: Variance
        \hat{v}_{\bm{s}_{t+1}} = 0
    \end{align}
\end{enumerate}

     % Complete mode
    \mypar{Complete} This is the default variant of DiSProD which uses all the variance terms. 
    As discussed above, for successful propagation of distribution, DiSProD requires at least some of the $J$ or $\tilde{H}$ terms to be non-zero.
\section{Experiment details}
\subsection{Environments}
In this section, we briefly describe the environments used for our experiments. The basic environments are taken from OpenAI Gym to which we make minor modifications. We rewrite the transition and reward functions by replacing any step-functions with their smooth equivalents. As the environments are deterministic, we add noise explicitly to make them stochastic.  $\epsilon$ denotes noise sampled from a standard Gaussian and $\alpha$  controls the amount of noise being added.
\subsubsection{Cartpole}
\label{sec: env-desc-cartpole}
Cartpole has four state variables - $x, y, \theta, \dot{\theta}$ and one control variable which controls the force acting on the cart. We make the environment stochastic by adding Gaussian noise to the force being applied. Concretely, the updated equation for force becomes $\text{force}_{\text{noisy}} = \text{force} + \alpha \epsilon$.
\subsubsection{Mountain Car}
\label{sec: env-desc-mountain-car}
Mountain Car has two state variables - position $(x)$ and velocity $(v)$ and one control variable ($u$) which is the force to be applied on the car. At a high level, the environment has two functions $f_x$ and $f_v$ that are used to update the position and velocity.
$$v_{t+1} = f_v(x_t, v_t) \text{ and } x_{t+1} = f_x(x_t, v_{t+1})$$
We add noise to $v_{t+1}$ to make the environment stochastic. $v_{t+1} = f_v(x_t, v_t) + \alpha \epsilon$.
\subsubsection{Pendulum}
Pendulum also has two state variables - $\theta$ and $\dot{\theta}$ and the control variable $(u)$ is the torque. First, $\dot{\theta}_{t+1}$ is computed using $\theta_t$, $\dot{\theta}_t$, $u_t$ and some constants, and then $\theta_{t+1}$ is updated using $\dot{\theta}_{t+1}$. We augment this and add noise to the update equation for $\theta_{t+1}$ such that it becomes $\theta_{t+1} = \theta_{t} + (\dot{\theta}_{t+1} + \alpha \exp(\epsilon)) dt$.

\begin{figure}
      \centering%
      \subfloat[\texttt{no-ob-1}]{%
        \resizebox{0.25\linewidth}{!}{%
          \begin{tikzpicture}%
            \node(a){\includegraphics[width=\linewidth]{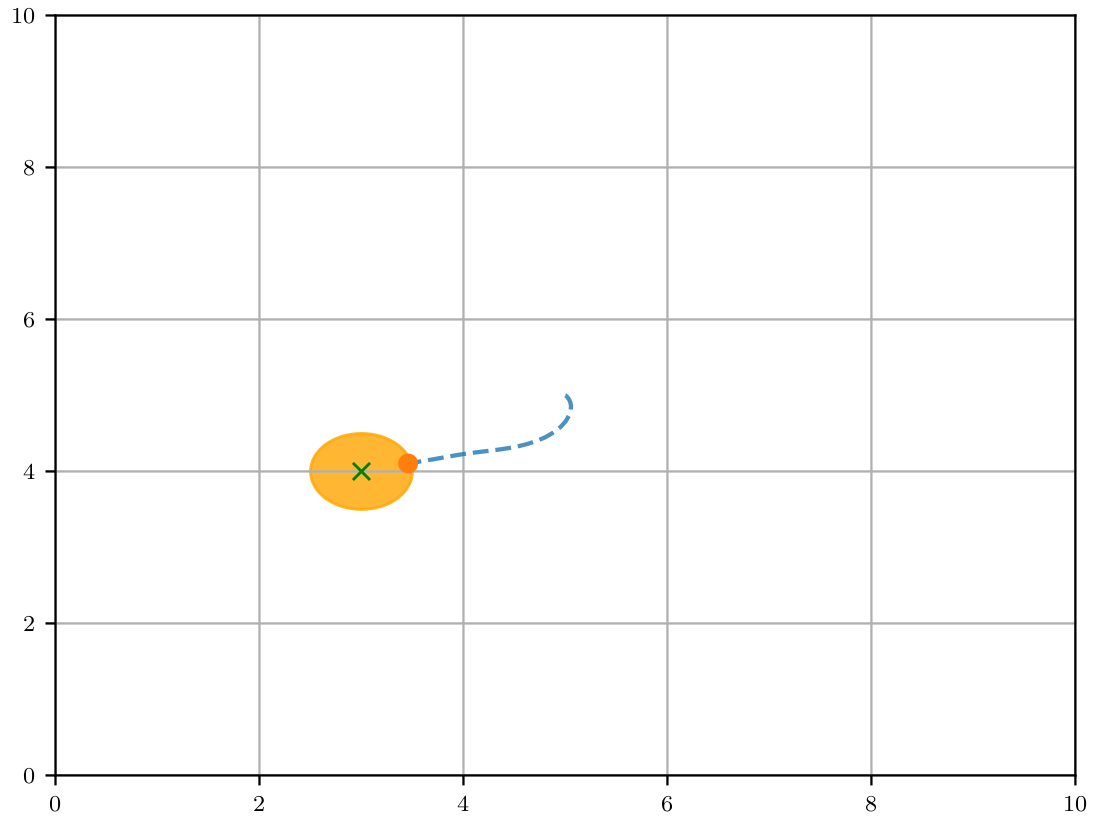}};%
          \end{tikzpicture}%
        }%
      }%
       \subfloat[\texttt{no-ob-2}]{%
        \resizebox{0.25\linewidth}{!}{%
          \begin{tikzpicture}%
            \node(a){\includegraphics[width=\linewidth]{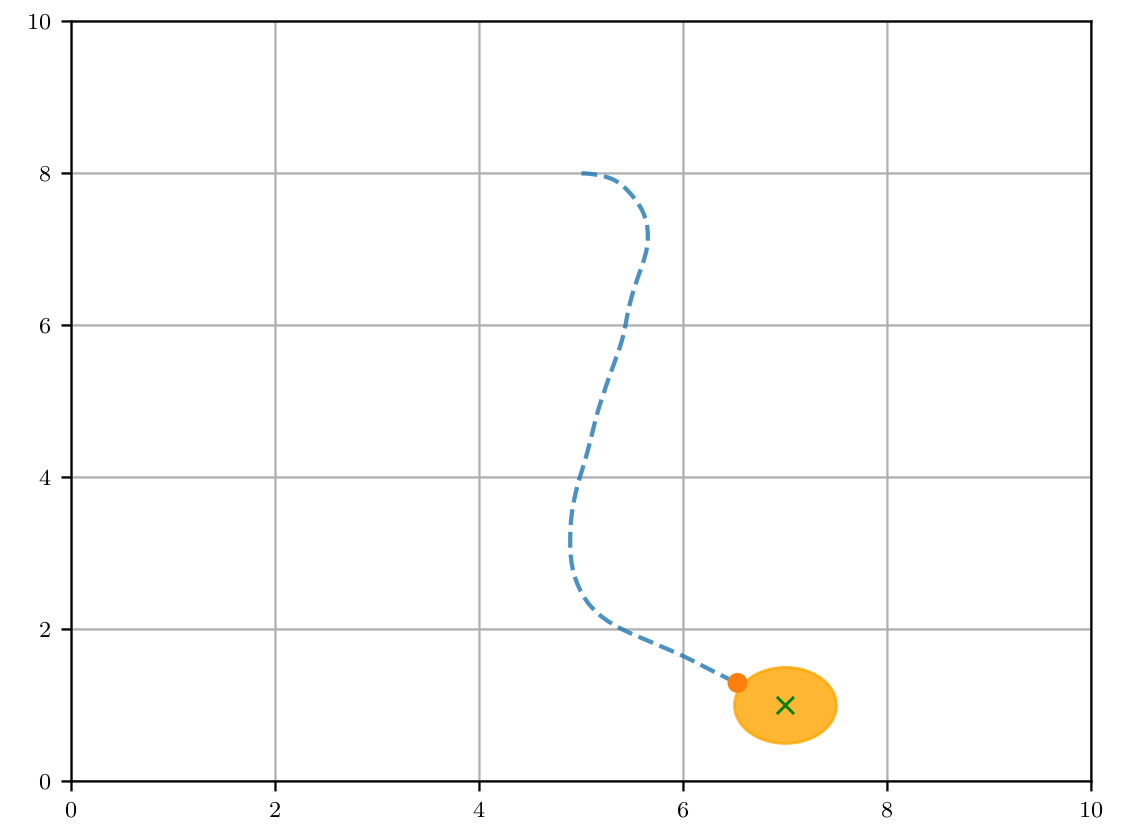}};%
          \end{tikzpicture}%
        }%
      }%
       \subfloat[\texttt{no-ob-3}]{%
        \resizebox{0.25\linewidth}{!}{%
          \begin{tikzpicture}%
            \node(a){\includegraphics[width=\linewidth]{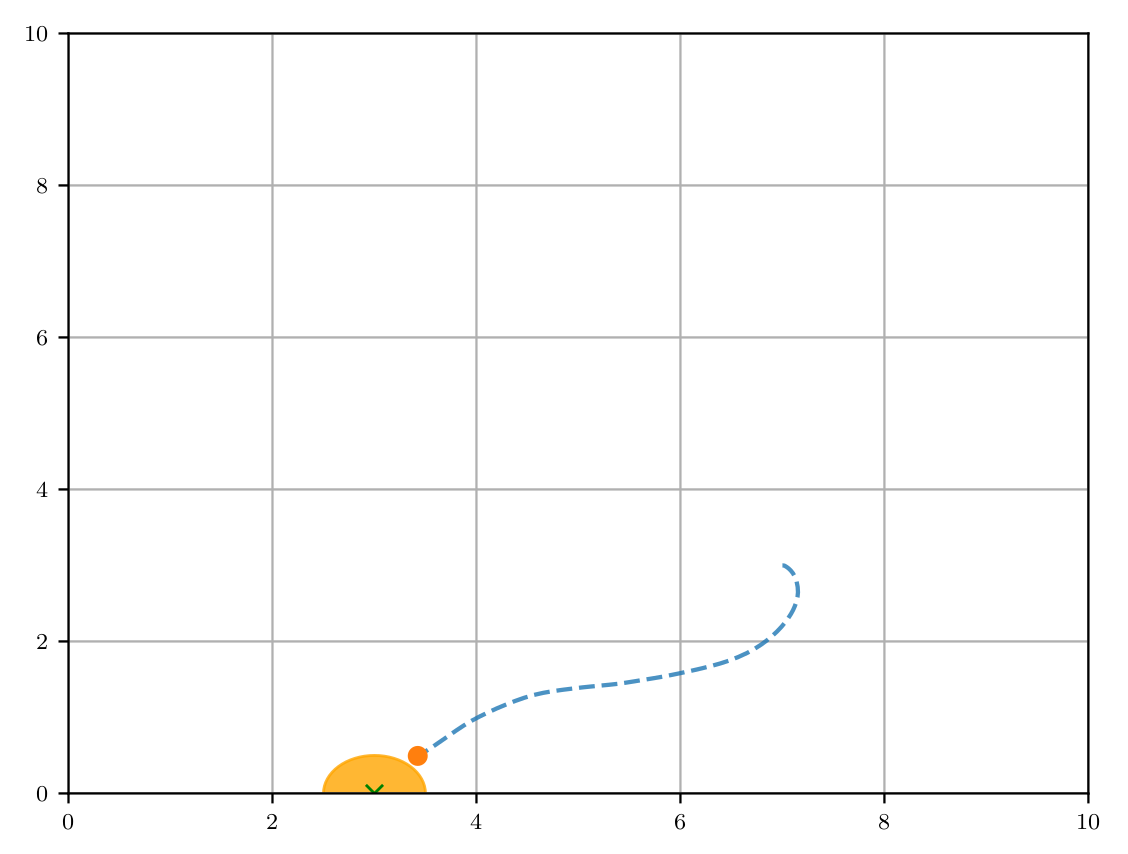}};%
          \end{tikzpicture}%
        }%
      }%
      \subfloat[\texttt{no-ob-4}]{%
        \resizebox{0.25\linewidth}{!}{%
          \begin{tikzpicture}%
            \node(a){\includegraphics[width=\linewidth]{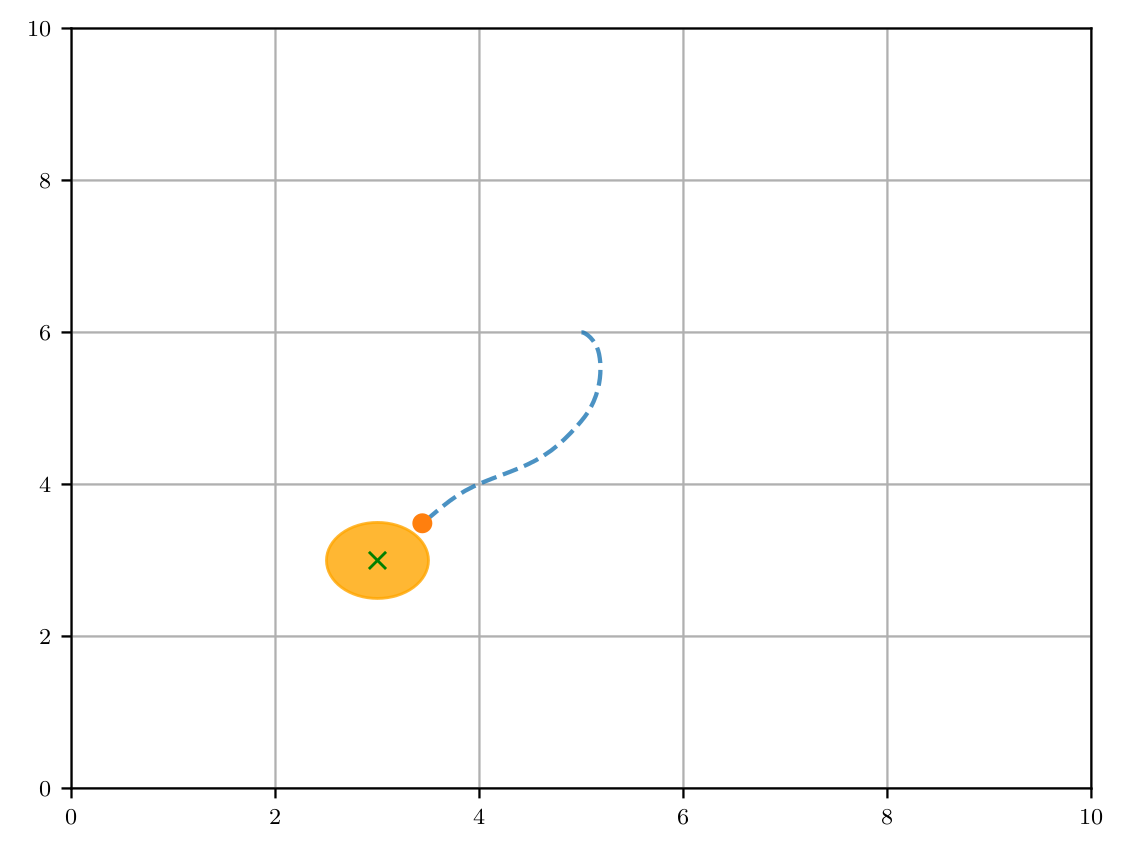}};%
          \end{tikzpicture}%
        }%
      }%
      \\
       \subfloat[\texttt{no-ob-5}]{%
        \resizebox{0.25\linewidth}{!}{%
          \begin{tikzpicture}%
            \node(a){\includegraphics[width=\linewidth]{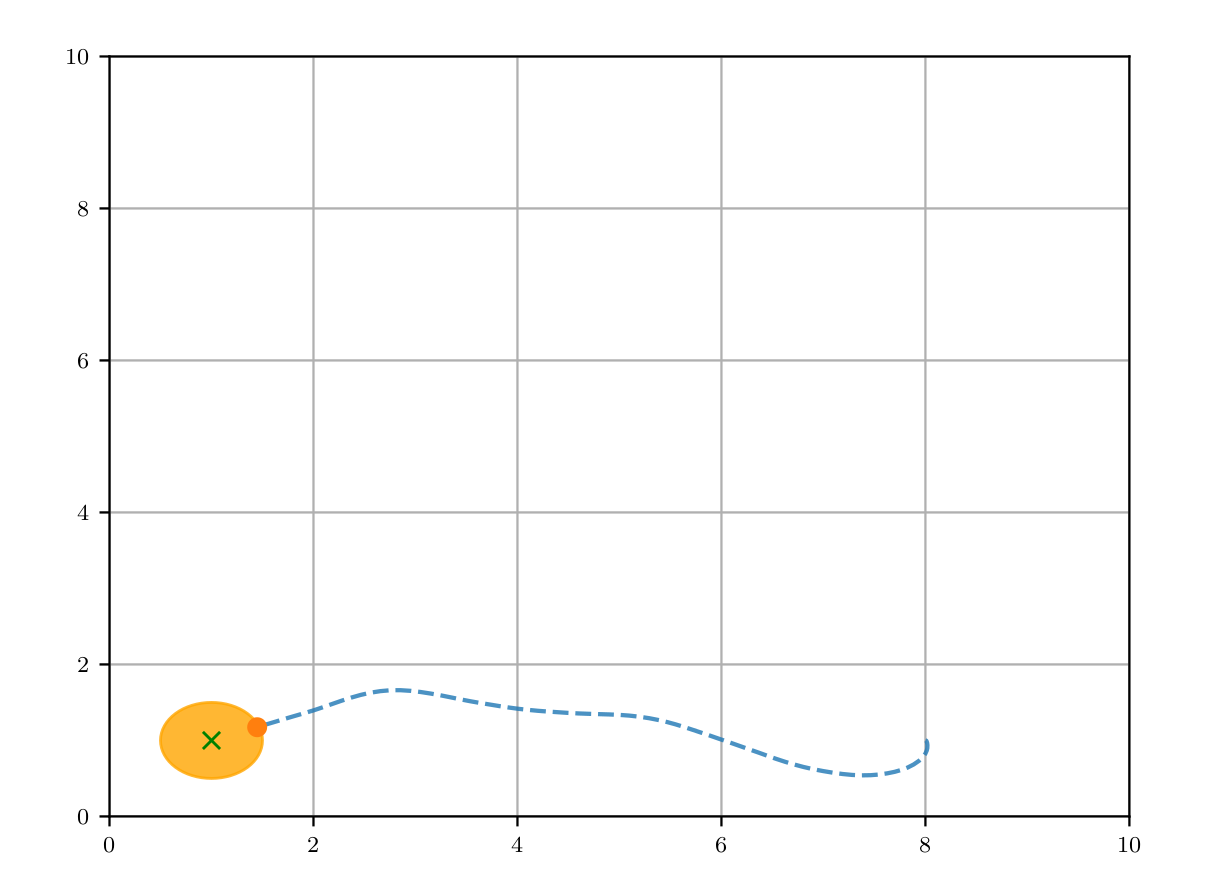}};%
          \end{tikzpicture}%
        }%
      }%
       \subfloat[\texttt{ob-1}]{%
        \resizebox{0.25\linewidth}{!}{%
          \begin{tikzpicture}%
            \node(a){\includegraphics[width=\linewidth]{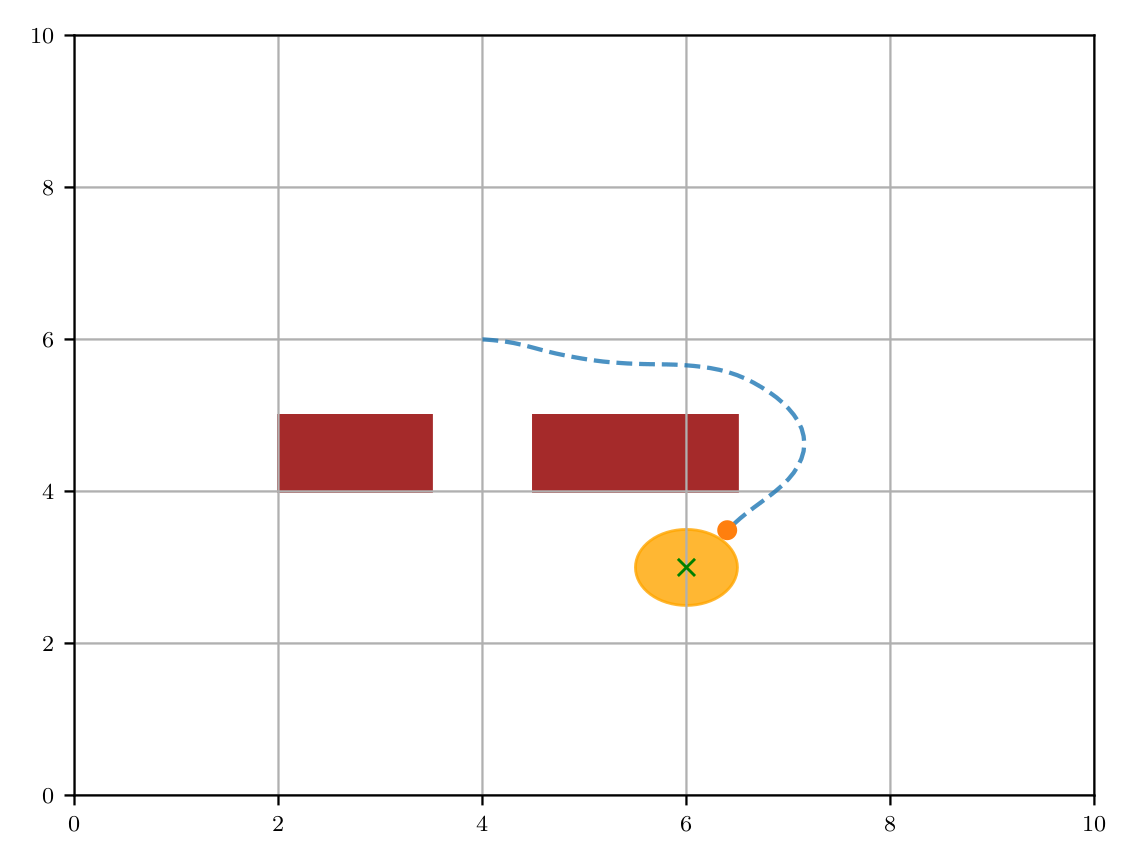}};%
          \end{tikzpicture}%
        }%
      }%
      \subfloat[\texttt{ob-2}]{%
        \resizebox{0.25\linewidth}{!}{%
          \begin{tikzpicture}%
            \node(a){\includegraphics[width=\linewidth]{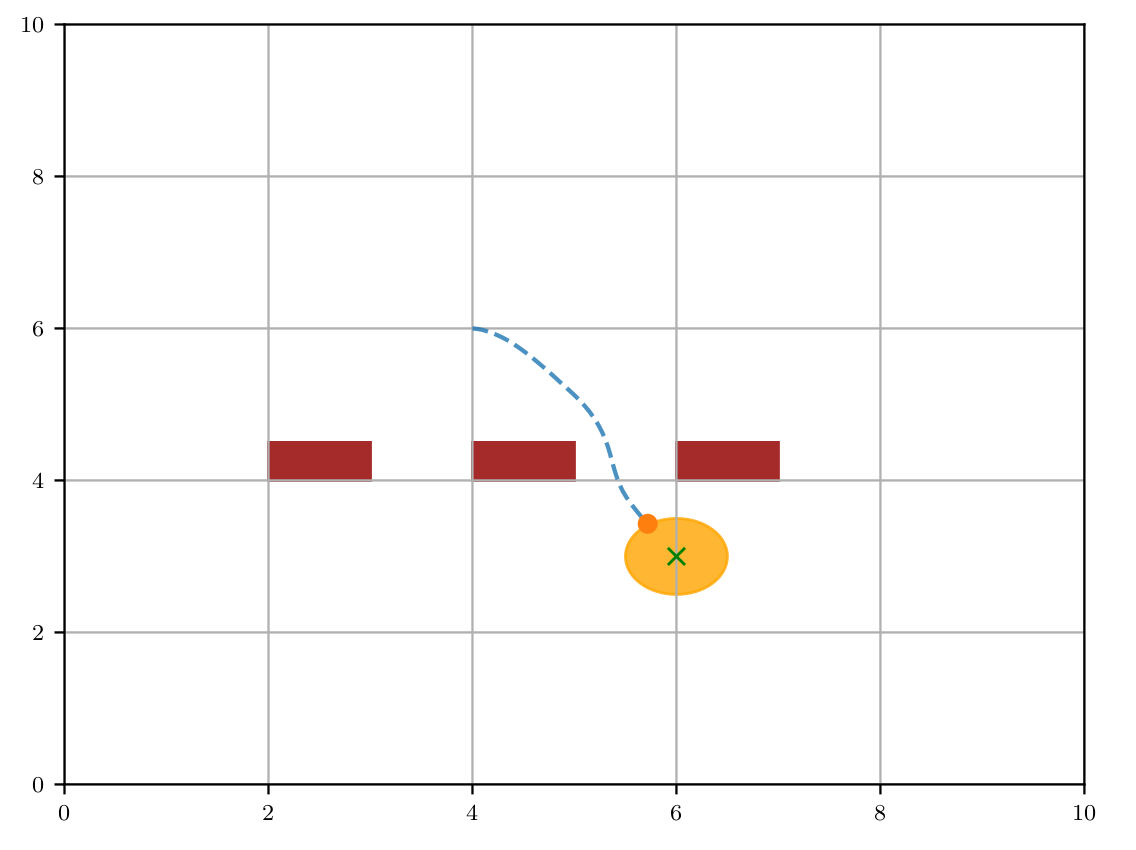}};%
          \end{tikzpicture}%
        }%
      }%
       \subfloat[\texttt{ob-3}]{%
        \resizebox{0.25\linewidth}{!}{%
          \begin{tikzpicture}%
            \node(a){\includegraphics[width=\linewidth]{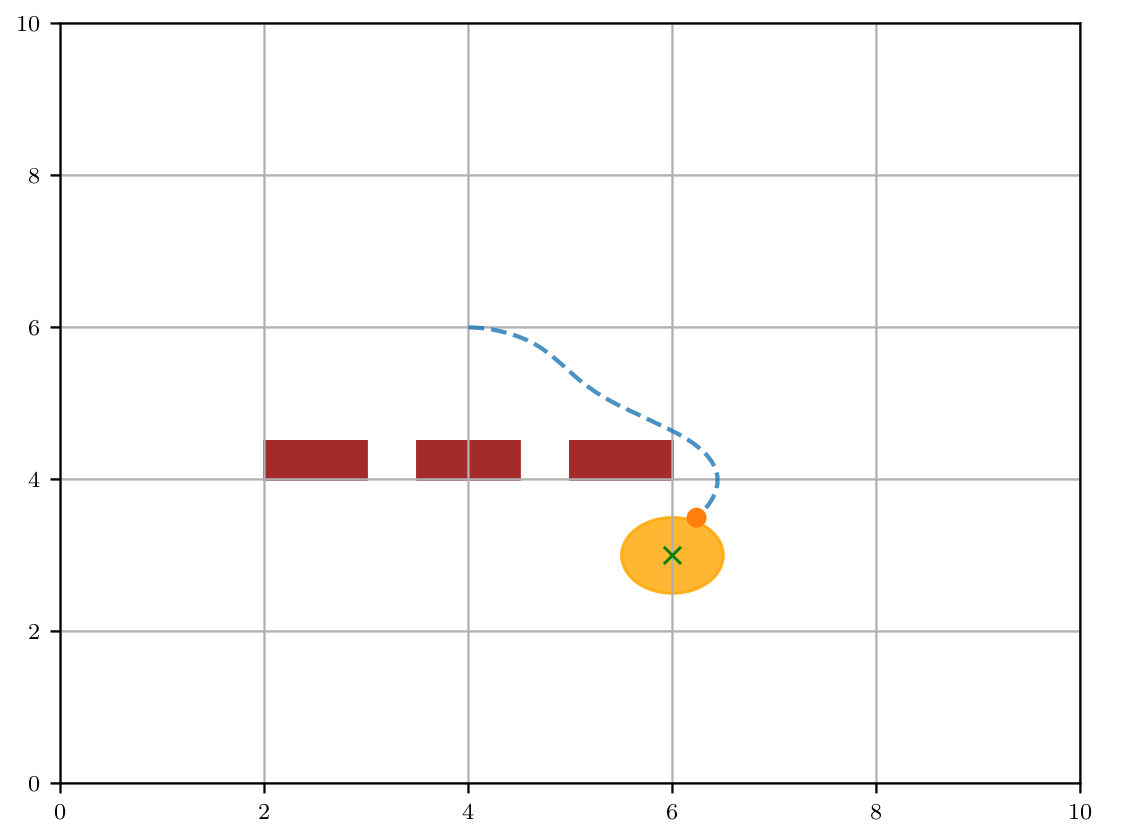}};%
          \end{tikzpicture}%
        }%
      }%
      \\
       \subfloat[\texttt{ob-4}]{%
        \resizebox{0.25\linewidth}{!}{%
          \begin{tikzpicture}%
            \node(a){\includegraphics[width=\linewidth]{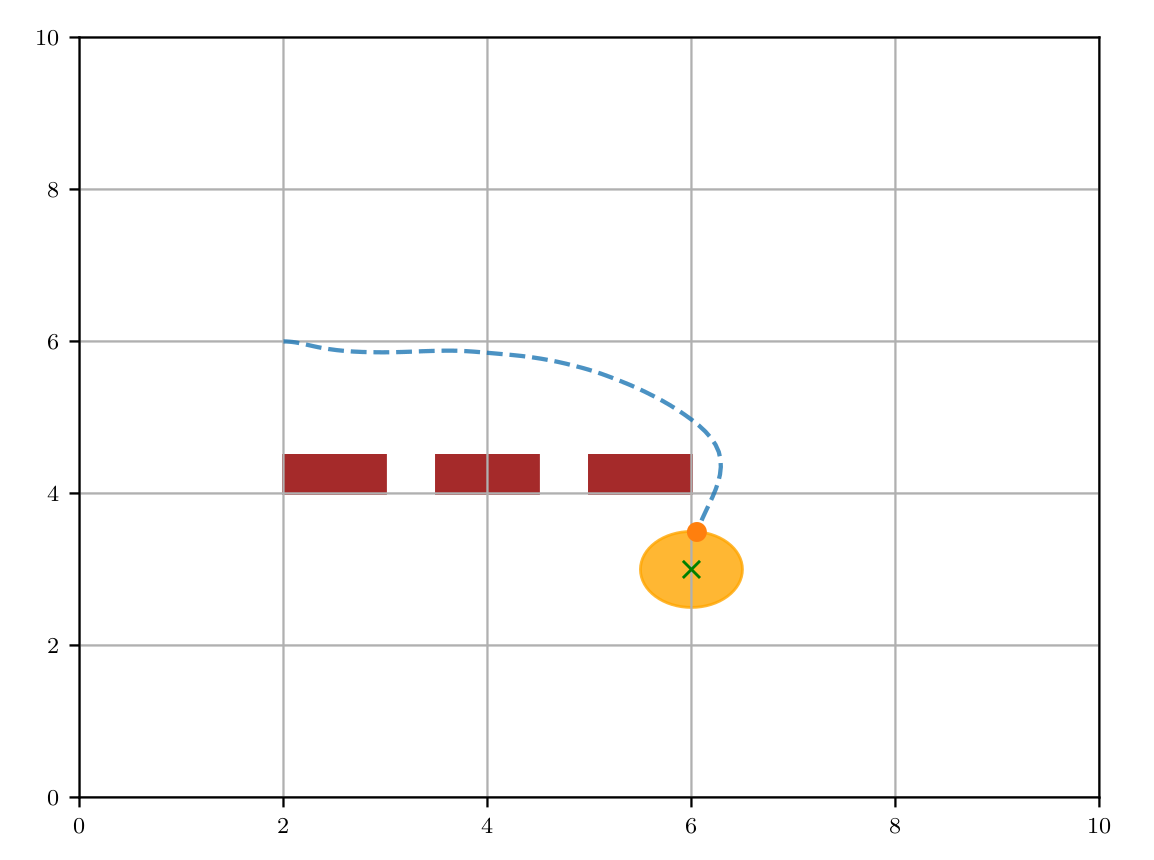}};%
          \end{tikzpicture}%
        }%
      }%
      \subfloat[\texttt{ob-6}]{%
        \resizebox{0.25\linewidth}{!}{%
          \begin{tikzpicture}%
            \node(a){\includegraphics[width=\linewidth]{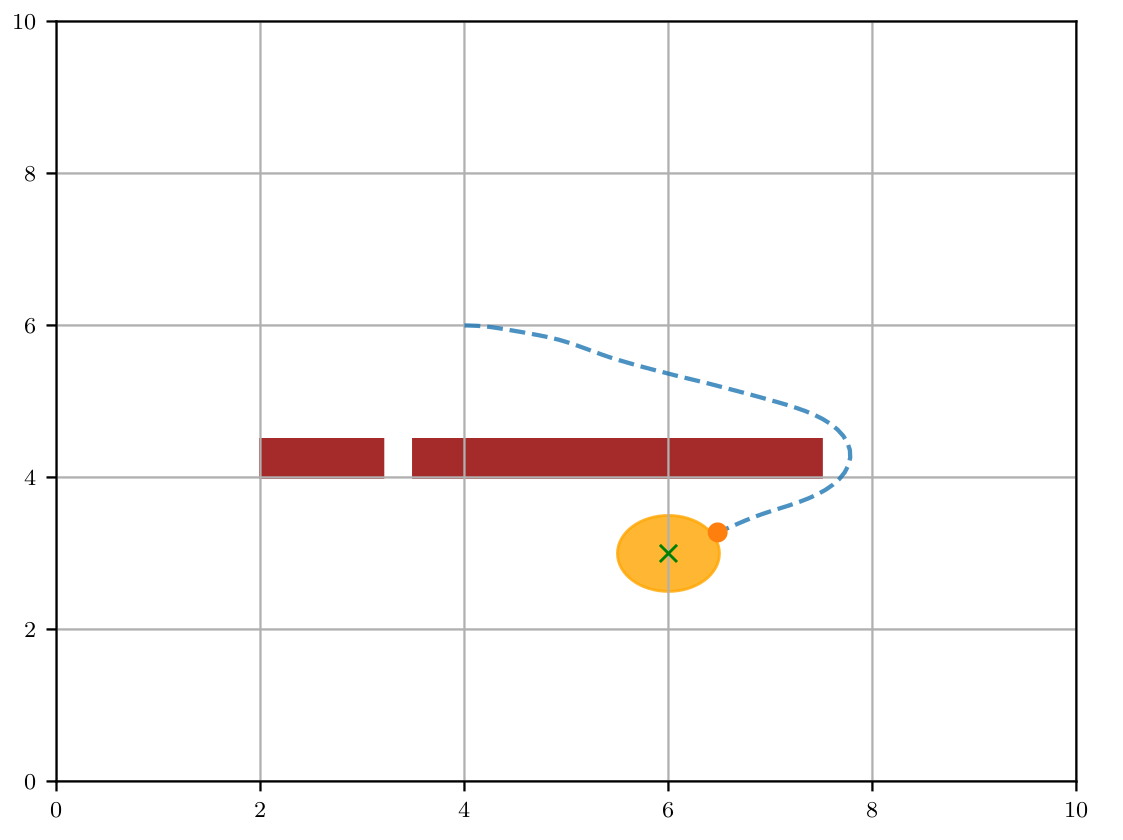}};%
          \end{tikzpicture}%
        }%
      }%
      \subfloat[\texttt{ob-7}]{%
        \resizebox{0.25\linewidth}{!}{%
          \begin{tikzpicture}%
            \node(a){\includegraphics[width=\linewidth]{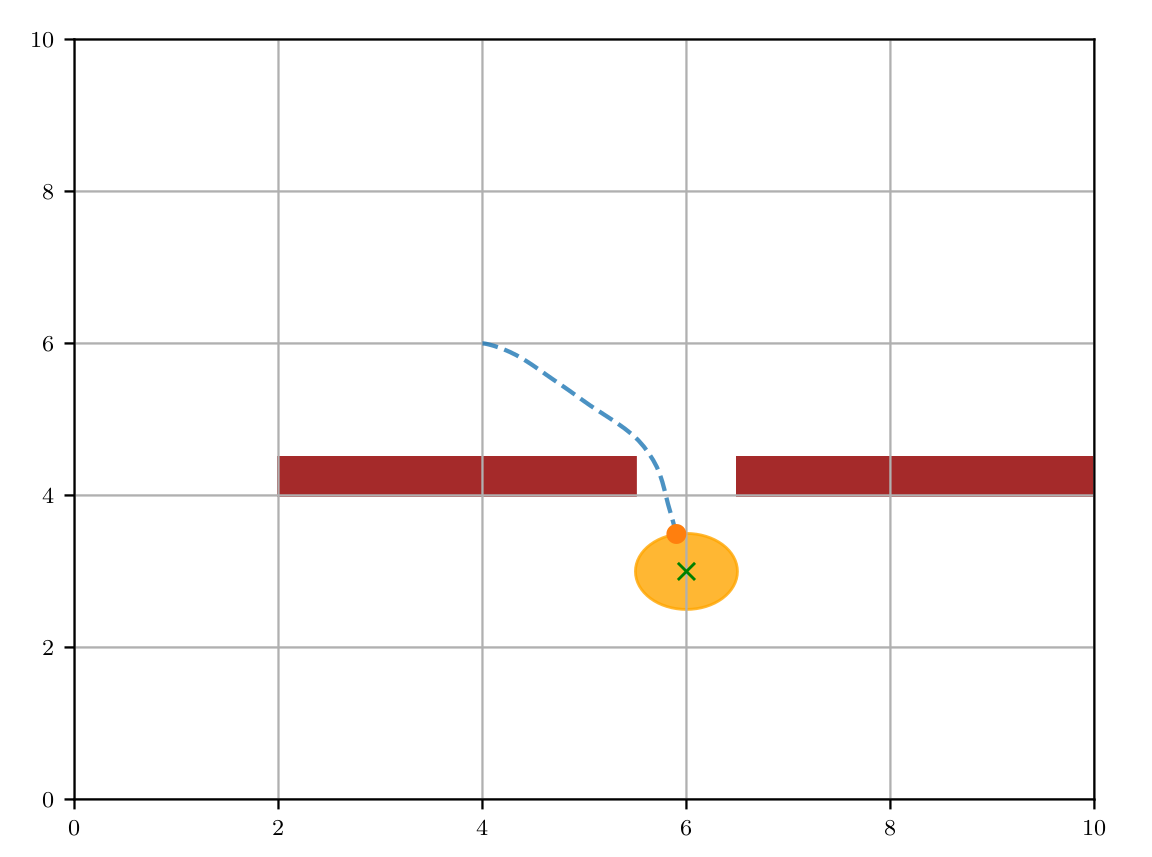}};%
          \end{tikzpicture}%
        }%
      }%
       \subfloat[\texttt{ob-8}]{%
        \resizebox{0.25\linewidth}{!}{%
          \begin{tikzpicture}%
            \node(a){\includegraphics[width=\linewidth]{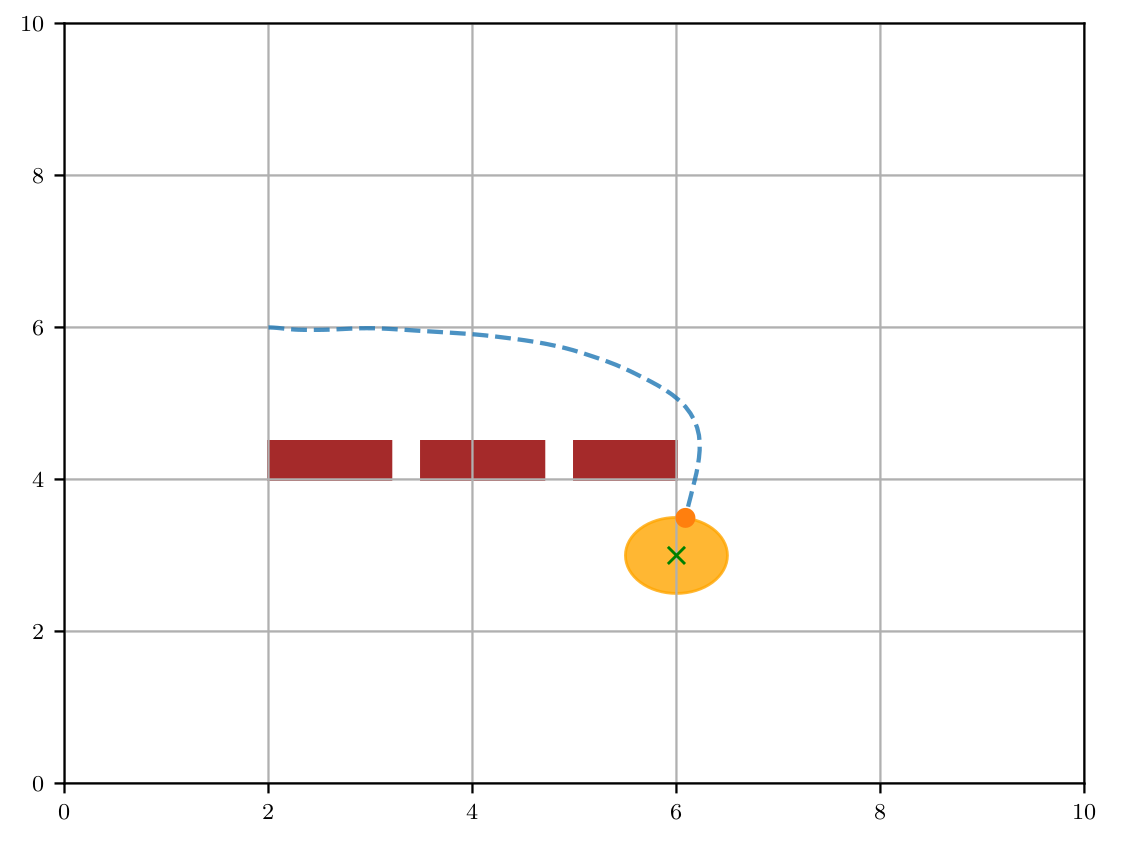}};%
          \end{tikzpicture}%
        }%
      }%
      \\
       \subfloat[\texttt{ob-9}]{%
        \resizebox{0.25\linewidth}{!}{%
          \begin{tikzpicture}%
            \node(a){\includegraphics[width=\linewidth]{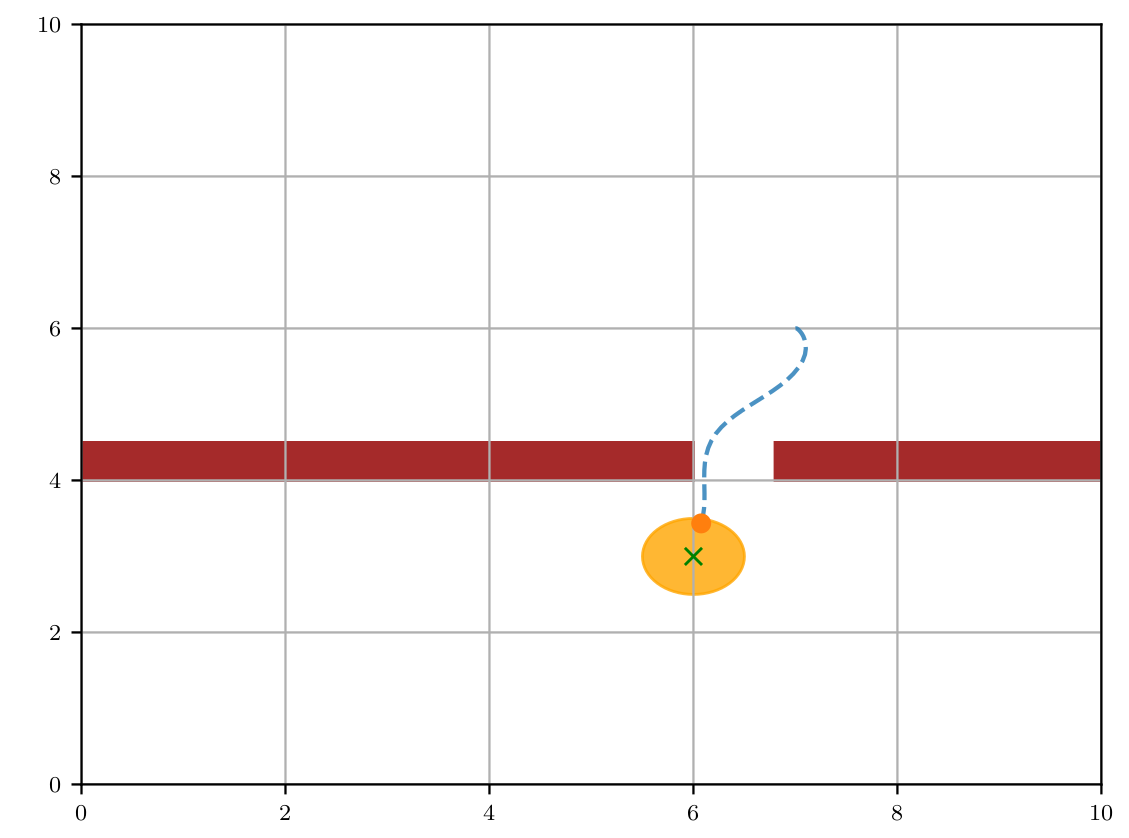}};%
          \end{tikzpicture}%
        }%
      }%
      \subfloat[\texttt{ob-10}]{%
        \resizebox{0.25\linewidth}{!}{%
          \begin{tikzpicture}%
            \node(a){\includegraphics[width=\linewidth]{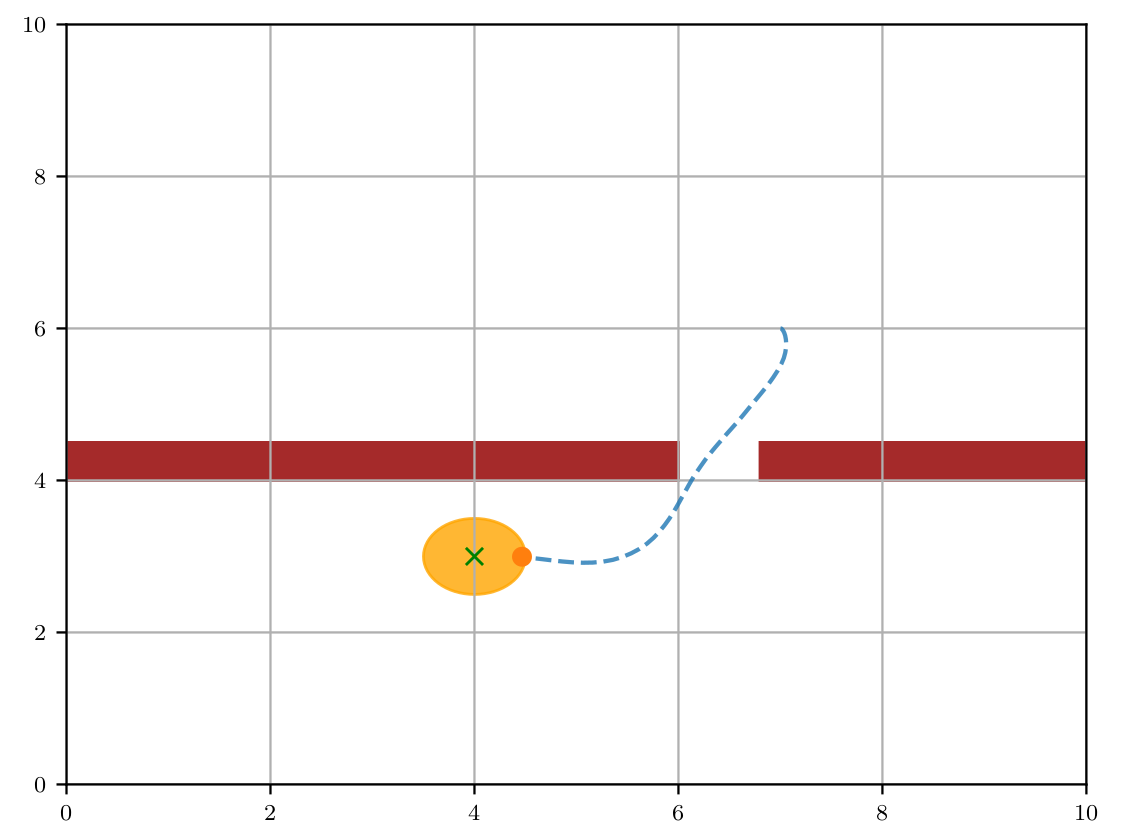}};%
          \end{tikzpicture}%
        }%
      }%
      \subfloat[\texttt{ob-11}]{%
        \resizebox{0.25\linewidth}{!}{%
          \begin{tikzpicture}%
            \node(a){\includegraphics[width=\linewidth]{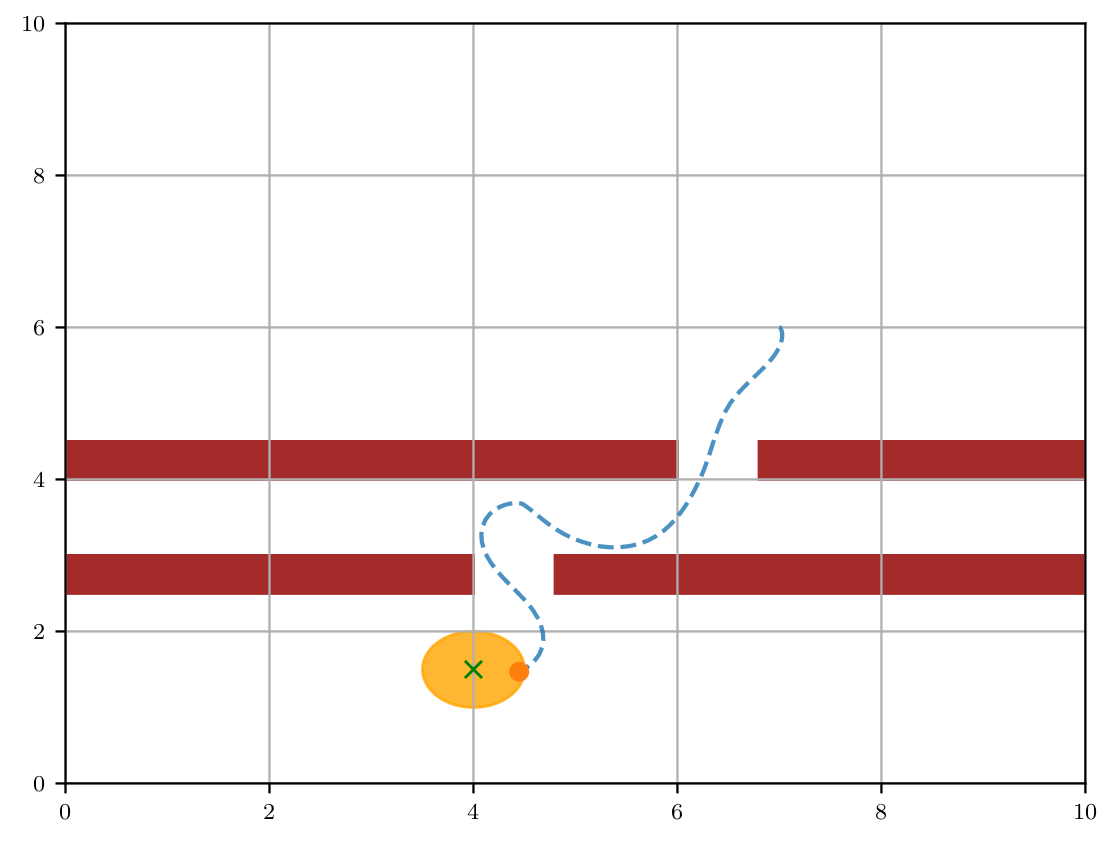}};%
          \end{tikzpicture}%
        }%
      }%
       \subfloat[\texttt{cave-mini}]{%
        \resizebox{0.25\linewidth}{!}{%
          \begin{tikzpicture}%
            \node(a){\includegraphics[width=\linewidth]{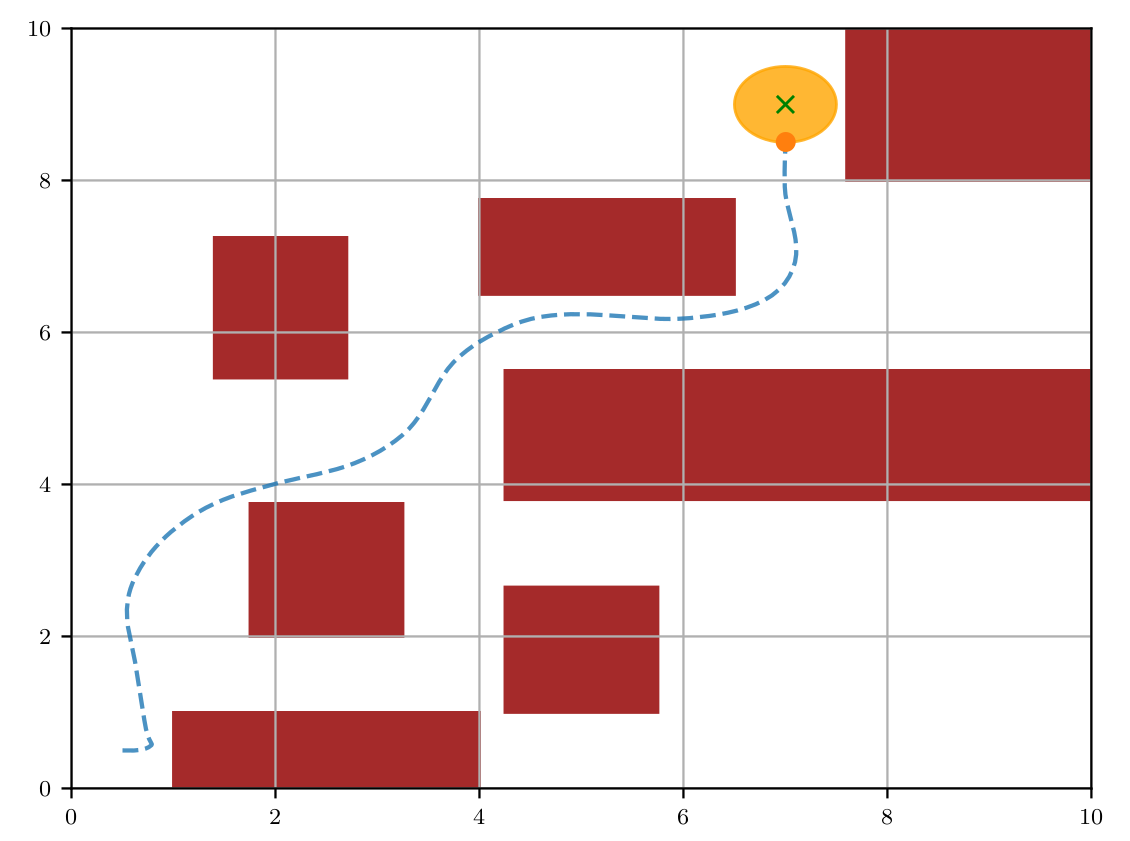}};%
          \end{tikzpicture}%
        }%
      }%
    \caption{Examples of maps on which experiments with Dubins' car in Gym were performed. Corresponding maps were generated for experiments in Gazebo.}\label{fig:Dubins car maps}
\end{figure}

\begin{figure}[th]%
  \centering%
 \subfloat[Map \texttt{ob-11} in RViz]{%
    \resizebox{0.45\linewidth}{!}{%
      \begin{tikzpicture}%
        \node(a){\includegraphics[width=\linewidth]{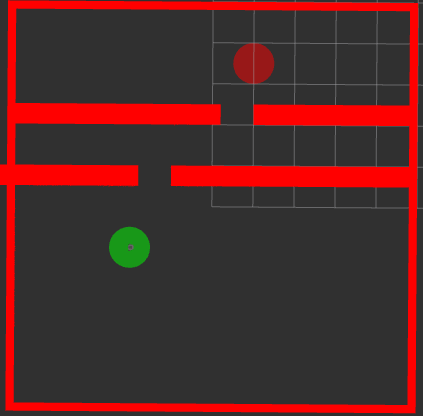}};%
      \end{tikzpicture}%
    }%
  }%
  \subfloat[Map \texttt{cave-mini} in RViz]{%
    \resizebox{0.45\linewidth}{!}{%
      \begin{tikzpicture}%
        \node(a){\includegraphics[width=\linewidth]{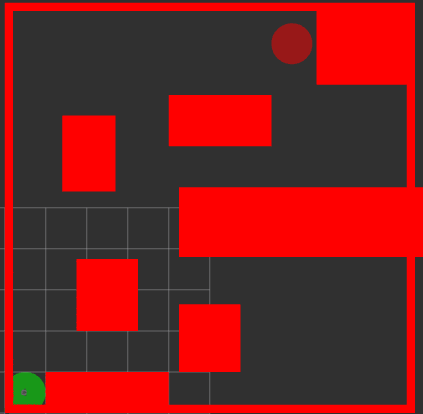}};%
      \end{tikzpicture}%
    }%
  }%
  \caption{Two of the maps used in simulated Turtelbot experiments in Gazebo, as visualized by RViz.}
  \label{fig:dubins}%
\end{figure}

\subsubsection{Dubins Car}
\label{sec: dubins_car_env_details}
The standard Dubins Car model maintains three state variables - $x, y, \theta$ and has two control variables - velocity ($v$) and angular velocity ($\omega$). The dynamics equations are:
\begin{align}
    x_{t+1} &= x_t + v_t \cos \theta_t dt \\
    y_{t+1} &= y_t + v_t \sin \theta_t dt \\
    \theta_{t+1} &= \theta_t + \omega dt
\end{align}
Since this is a discrete time system and ignores acceleration, the motion of the car can be unrealistic. For example, although $v_t = 0$ and $v_{t+1} = v_{\max}$ is a valid control, this cannot be performed by a real vehicle. As Gazebo is a real-physics simulator and considers factors like acceleration, inertia and friction, planning with this model leads to poor performance in Gazebo. In order to obtain a better approximation of the model used by Gazebo, we modify the controls from velocity and angular velocity to delta velocity $(\Delta v)$ and delta angular velocity $(\Delta \omega)$ and compute velocity and angular velocity as
\begin{align}
    v_t &= v_{t-1} + \Delta v_{t} \\
    \omega_t &= \omega_{t-1} + \Delta \omega_{t}
\end{align}
This prevents sudden changes in linear and angular velocity and helps to limit the actions. We experiment with this model on a suite of maps ranging from simple ones with no obstacles to complex ones with obstacles.
All the map configurations are shown in \Cref{fig:Dubins car maps} and two of the RViz maps are shown in \Cref{fig:dubins}.

\subsubsection{Simple Env}
\label{sec:simple_env_details}
In the Gym environments that we experimented with, some of the required partial derivatives were either zero or not significant enough to make a difference in the planning results. To showcase the difference more clearly, we designed an additional environment called SimpleEnv where all the required partials are non-zero.

The environment has two state variables - $x$ and $y$ and two action variables - $\delta_x$ and $\delta_y$.
The dynamics are as follows:
\begin{align}
    x_{t+1} &= x + \delta_x + \alpha (0.1\epsilon + \epsilon^2) \\
    y_{t+1} &= y + \delta_y
\end{align}
The reward function is a smoothed 0-1 reward where a reward of 1 is given if the agent is sufficiently close to a predefined goal position.

\subsubsection{Cartpole - Hybrid Variant}
To experiment with hybrid environments, we modify Cartpole and introduce an environment variable called \textit{reward marker} (set to 0.5 in the experiment) and a new binary state variable that is set to 1 if the cart is to the left of the reward marker. The reward function is modified such that the agent gets a reward of 3 if the agent is to right of the reward marker and otherwise it continues to get a reward of 1. The initial position of the cart is always to the left of the reward marker. The noise is added in a similar manner as in the original Cartpole environment. 

\subsubsection{Mountain Car - High Dimensional Action Space}
For our experiments with high dimensional action space, we tweak the standard Mountain Car environment 
%detailed in \Cref{sec: env-desc-mountain-car} 
and add $m$ redundant action variables. Similar to the original Mountain Car environment, the dynamics are only influenced by the 1st action variable. We modify the action penalty term in the reward function to sum over all the action variables. Specifically, the modified action penalty term function becomes $0.1 \sum_{i=1}^m a_i$.

\subsection{Controlling Sparsity of Rewards}
\begin{figure}
      \centering
    \resizebox{0.75\linewidth}{!}{%
      \begin{tikzpicture}%
         \node{\includegraphics[width=\linewidth]{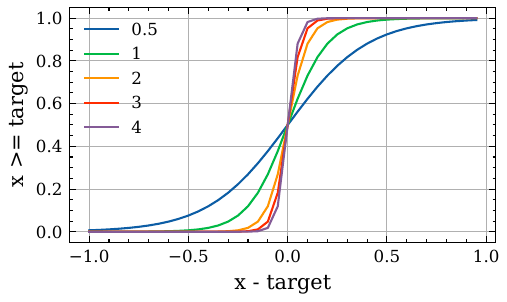}};%
      \end{tikzpicture}%
    }%
    \caption{As the value of sparsity multiplier increases, the greater-than-equal-to grows closer to a step function.}
    \label{fig:smooth_greater_than}
\end{figure}

As mentioned in \Cref{sec:planning_with_sparse_rewards}, we modify the default reward function of Continuous Mountain Car to use a smooth version of greater-than-equal-to function $(f_{ge})$. The function takes as input a variable $x$ and a target $t$ and outputs $\sigma(10\beta(x-t))$ where $\beta$ is the sparsity multiplier and $\sigma(a) = \nicefrac{1}{1+e^{-a}}$. The function is a smooth approximation of $x \ge t$.
 \Cref{fig:smooth_greater_than} shows a plot of $f_{ge}$ against $x-t$.
As the sparsity multiplier increases, the value of the function becomes increasingly similar to that of a step function and rewards become sparse.

\subsection{Hardware Platforms}
We use two physical robots in our experiments - Jackal and Heron, both of which are made by Clearpath Robotics. As mentioned in \Cref{sec:evaluation_with_a_physics_simulator}, in both the cases, the positions of obstacles in the environment are unknown to the transition model. However, the reward function is aware of the obstacles and penalizes trajectories that collide with one.

Jackal is an Unmanned Ground Vehicle~(UGV), whose position and orientation are obtained through a VICON Motion Capture System, and the UGV is controlled directly by \disprodSpace through commands for linear and angular velocity. 

Heron is an Unmanned Surface Vessel~(USV), whose position is obtained by fusing GPS and Inertial Measurement Unit~(IMU) information using EKF.
It is also controlled through commands for linear and angular velocity, but in our experiments this is mediated through a PID controller because one of the USV's thrusters has intermittent motor failures, which requires a high-frequency feedback controller to compensate the significant error.
\disprodSpace computes an action sequence through its computation graph and then send a list of waypoints -- the means of states on the graph -- to the PID controller.
The list is updated asynchronously at every decision cycle of the planner. 

\subsection{Additional Details}
We list the values of the parameters used for our experiments with OpenAI Gym in \Cref{tab:gym_param_specs} and for our experiments with Jackal and Heron in \Cref{tab:hardware_param_specs}. The step sizes for $\mu_{\bm{a}}$ and $v_{\bm{a}}$ are indicated by $lr_\mu$ and $lr_v$. $\overline{\Delta v}$ and $\overline{\Delta \omega}$ represent the maximum permissible values for delta linear velocity and delta angular velocity, while $\overline{v}$ and $\overline{\omega}$ indicate the maximum permissible linear and angular velocity respectively. 

\begin{table}[htbp]
  \centering
  \scriptsize
  \begin{tabular}{lcccccc}
    \toprule
    {Environments} & {nA} & {nS} & {$lr_\mu$} & {$lr_v$} & {Rollout Depth} & {Restarts}\\
    \midrule
    CartPole & 1 & 4 & 10 & 1 & 25 & 200 \\
    Pendulum & 1 & 4 & 1 & 0.1 & 25 & 200\\
    Mountain Car & 1 & 2 & 0.1 & 0.001 & 100 & 200 \\ 
    Dubins Car & 5 & 5 & 10 & 1 & 100 & 200\\    
    Simple Env & 2 & 2 & 0.01 & 0.001 & 20 & 50 \\
    \bottomrule
  \end{tabular}
  \caption{Parameter specifications for experiments with OpenAI Gym simulators.}\label{tab:gym_param_specs}
\end{table}

\begin{table}[htbp]
  \centering
  \scriptsize
  \begin{tabular}{lcccccc}
    \toprule
    {Robots} & {$\overline{\Delta v}$} & {$\overline{\Delta \omega}$} & {$\overline{v}$} & {$\overline{\omega}$} &  {Rollout Depth} & {Restarts}\\
    \midrule
    TurtleBot  & 0.05 & 1 & 0.5 & 60 & 100 & 200 \\
    Jackal  & 0.3 & 5 & 0.6 & 60 & 30 & 400\\
    Heron  & 0.5 & 10 & 1 & 30 & 70 & 400  \\        
    \bottomrule
  \end{tabular}
  \caption{This table elaborates the parameters of the Dubin's car model with action ranges, and specifies modifications to the depth and restarts in experiments with TurtleBot, Jackal and Heron.}\label{tab:hardware_param_specs}
\end{table}
\section{Experimental Results}
% Planning horizon
\begin{figure*}[htbp]%
  \centering%
  \subfloat[CCP $(\alpha=0)$\label{fig:exp_depth_cartpole}]{%
    \resizebox{0.3\linewidth}{!}{%
      \begin{tikzpicture}%
        \node(a){\includegraphics[width=\linewidth]{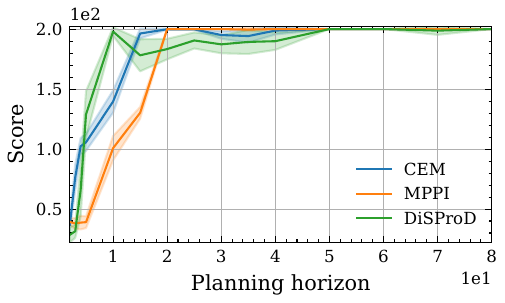}};%
      \end{tikzpicture}%
    }%
  }%
  \subfloat[P $(\alpha=0)$\label{fig:exp_depth_pendulum}]{%
    \resizebox{0.3\linewidth}{!}{%
      \begin{tikzpicture}%
        \node(a){\includegraphics[width=\linewidth]{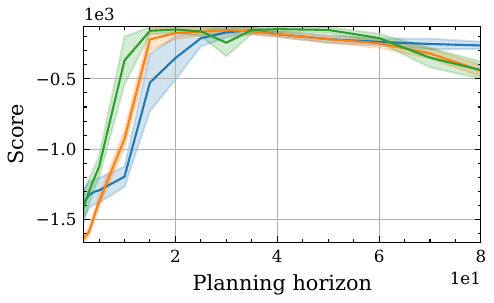}};%
      \end{tikzpicture}%
    }%
  }%
  \subfloat[CMC $(\alpha=0)$\label{fig:exp_depth_mountain_car}]{%
    \resizebox{0.3\linewidth}{!}{%
      \begin{tikzpicture}%
        \node(a){\includegraphics[width=\linewidth]{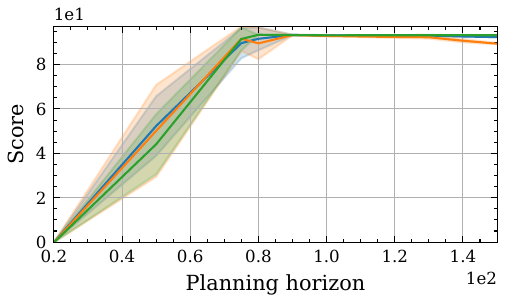}};%
      \end{tikzpicture}%
    }%
  }%
\\
 \subfloat[CCP $(\alpha=0)$\label{fig:exp_restarts_cartpole}]{%
    \resizebox{0.3\linewidth}{!}{%
      \begin{tikzpicture}%
        \node(a){\includegraphics[width=\linewidth]{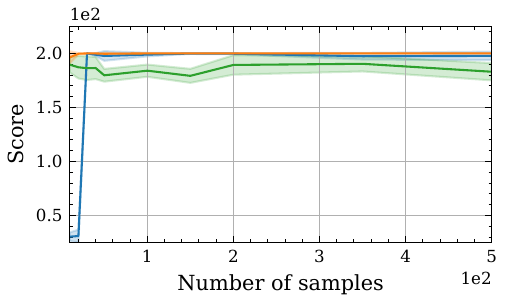}};%
      \end{tikzpicture}%
    }%
  }%
  \subfloat[P  $(\alpha=0)$\label{fig:exp_restarts_pendulum}]{%
    \resizebox{0.3\linewidth}{!}{%
      \begin{tikzpicture}%
        \node(a){\includegraphics[width=\linewidth]{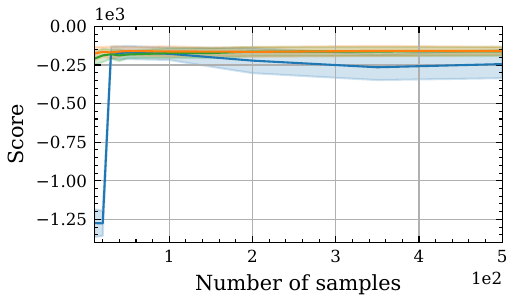}};%
      \end{tikzpicture}%
    }%
  }%
  \subfloat[CMC  $(\alpha=0)$\label{fig:exp_restarts_mountain_car}]{%
    \resizebox{0.3\linewidth}{!}{%
      \begin{tikzpicture}%
        \node(a){\includegraphics[width=\linewidth]{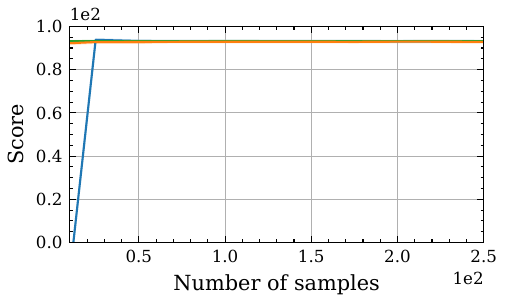}};%
      \end{tikzpicture}%
    }%
  }%
\\
  \centering%
  \subfloat[CCP  $(\alpha=5)$\label{fig:exp_restarts_cartpole_noisy}]{%
    \resizebox{0.3\linewidth}{!}{%
      \begin{tikzpicture}%
        \node(a){\includegraphics[width=\linewidth]{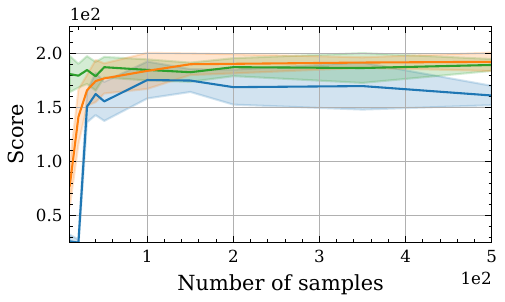}};%
      \end{tikzpicture}%
    }%
  }%
  \subfloat[P  $(\alpha=2)$\label{fig:exp_restarts_pendulum_noisy}]{%
    \resizebox{0.3\linewidth}{!}{%
      \begin{tikzpicture}%
        \node(a){\includegraphics[width=\linewidth]{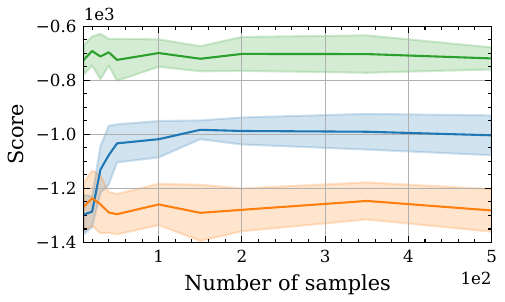}};%
      \end{tikzpicture}%
    }%
  }%
  \subfloat[CMC  $(\alpha=0.002)$\label{fig:exp_restarts_mountain_car_noisy}]{%
    \resizebox{0.3\linewidth}{!}{%
      \begin{tikzpicture}%
        \node(a){\includegraphics[width=\linewidth]{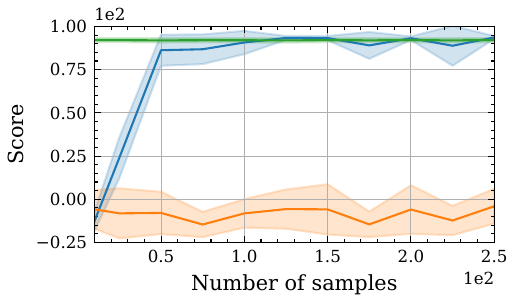}};%
      \end{tikzpicture}%
    }%
  }%
  \caption{Environments used are Continuous Cartpole (CCP), Pendulum (P) and Continuous Mountain Car (CMC). \ref{fig:exp_depth_cartpole}, \ref{fig:exp_depth_pendulum}, \ref{fig:exp_depth_mountain_car}: In deterministic environments $(\alpha = 0)$, optimal performance is achieved with a small planning horizon. Increasing the planning horizon further does not impact performance of any of the planners. \ref{fig:exp_restarts_cartpole}, \ref{fig:exp_restarts_pendulum}, \ref{fig:exp_restarts_mountain_car}: A similar behaviour is observed when number of samples is varied and $\alpha = 0$.  \ref{fig:exp_restarts_cartpole_noisy}, \ref{fig:exp_restarts_pendulum_noisy}, \ref{fig:exp_restarts_mountain_car_noisy}: In noisy environments, DiSProD performs better than CEM and MPPI. Interestingly, increasing the number of samples does not improve the performance. Note that the hyperparameters used are same in both deterministic and stochastic environments.
  }\label{fig:exp_appendix_gym_results}%\vspace{-20pt}%
\end{figure*}

\myparquestion{Does having a larger planning horizon help in deterministic environments?}
In our experiments with basic Gym environments, we observe that optimal performance, with all the three planners, is achieved with a small planning horizon. Increasing the planning horizon further does not impact the performance. (\Cref{fig:exp_appendix_gym_results}~(a-c)).

% Exp: Varying number of samples
\myparquestion{Does sampling more trajectories help?}
Shooting algorithms sample action sequences to generate trajectories. Intuitively, increasing the number of action sequences that a planner is allowed to sample should help with better plans and hence improve performance. 
We observe that in deterministic environments, sampling a small number of trajectories is sufficient to obtain near optimal performance. Making the environments stochastic degrades the planners' overall performance. Interestingly enough, in these environments, the performance remains unaffected on increasing the number of samples. Experiments with high dimensional Mountain Car indicate that this behaviour might be due to the fact the state and action space for the basic Gym environments is quite small.  Note that we use the same hyperparameters in deterministic and stochastic environments. 
(\Cref{fig:exp_appendix_gym_results}~(d-i)).

% Planning horizon
\begin{figure*}[ht]%
  \centering%
  \subfloat{%
    \resizebox{\linewidth}{!}{%
      \begin{tikzpicture}%
        \node(a){\includegraphics[width=\linewidth]{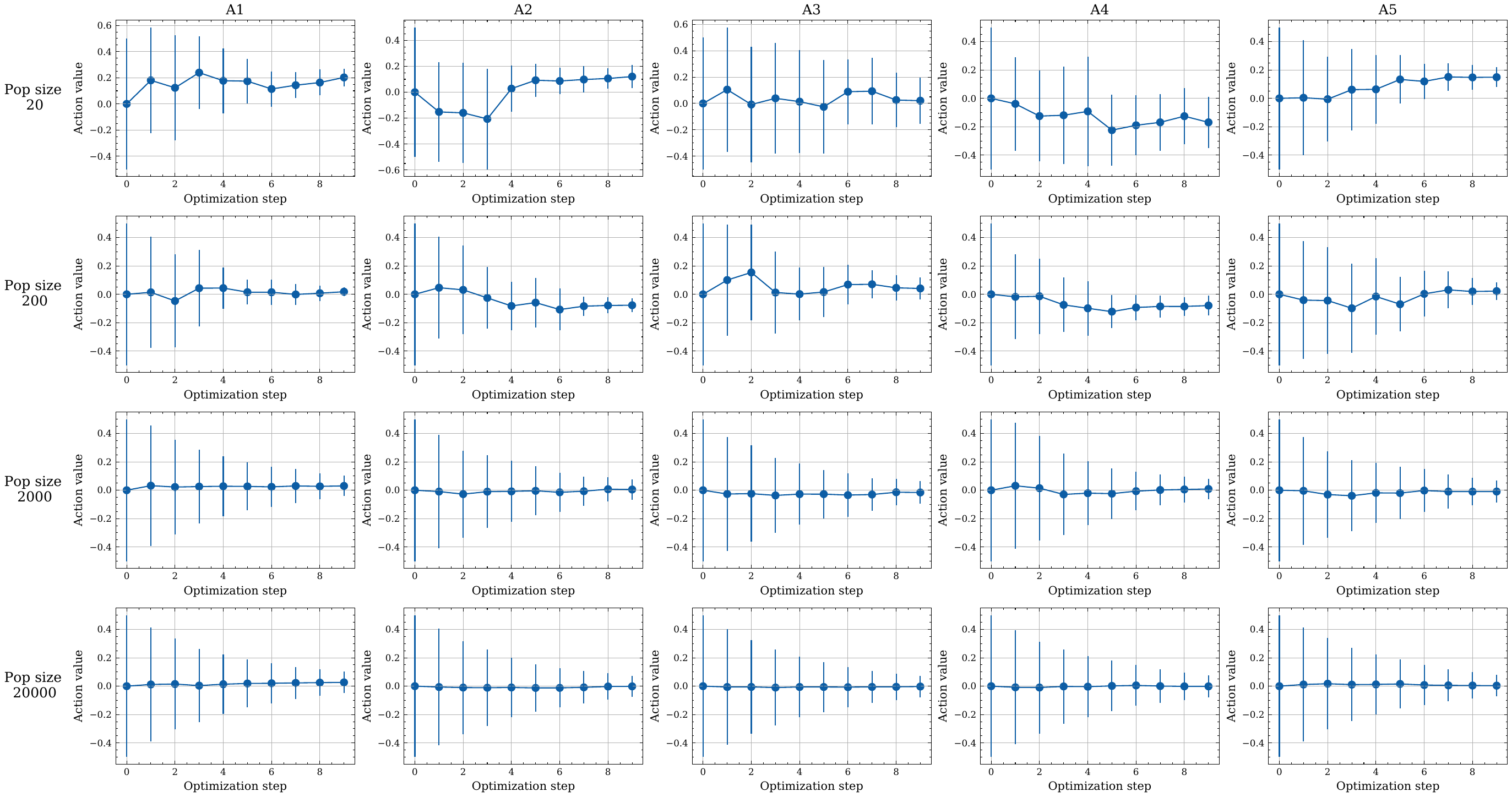}};%
      \end{tikzpicture}%
    }%
  }%
  \caption{Values of 5 action variables (A1 to A5) across 10 optimization steps as the population size is increased. The action value at the 10th optimization step is executed in the environment. Only A1 influences the dynamics, whereas all of the action variables contribute to action penalty in the reward function. An agent must learn to optimize A1 to achieve the goal, while keeping the values of A2 to A5 close to 0 and achieve a higher reward. When the population size is small, CEM fails to achieve this and obtains a poor score.}\label{fig:exp_high_dim_cem_convergence}%\vspace{-20pt}%
\end{figure*}
\myparquestion{How does the performance of shooting algorithms vary when the size of the action space increases?}
When the action space increases, a small number of samples cover a very small volume of the possible trajectories. Intuitively, the performance of agents relying on shooting methods will be poor in such scenarios and performance should improve on increasing the number of samples.  As shown in \Cref{fig:exp_high_dim_scores}, this indeed is the case.  We take a closer look at the performance of CEM as the population size increases. For this, we use the modified version of Mountain Car as detailed in \Cref{sec: experiment_with_high_dim_action_space} with 14 redundant actions. 
For this experiment we fix the state and consider action selection in CEM which is done with $K=10$ optimization steps. 
We plot the values of 5 action variables (A1 to A5) across 10 optimization steps where A1 influences the dynamics while the rest are redundant action variables. The action value after the 10th optimization step is output by CEM and is executed by the agent in the environment. For attaining high rewards, an agent must learn to keep the values of action variables A2 through A5 as close to 0 as possible, while using A1 to act optimally. 
The result is shown in \Cref{fig:exp_high_dim_cem_convergence}.
We observe that as the population size increases, CEM does a better job at keeping the values of the redundant variables close to 0. Plots shown in the main paper show that this indeed corresponds to improved performance.

\begin{figure*}[th]%
  \centering%
  \subfloat{%
        \includegraphics[width=0.4\linewidth]{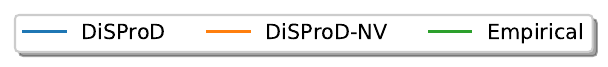}
    }\\
    \setcounter{subfigure}{0}
  \subfloat[$\alpha = 0$]{%
    \resizebox{0.25\linewidth}{!}{%
      \begin{tikzpicture}%
        \node(a){\includegraphics[width=\linewidth]{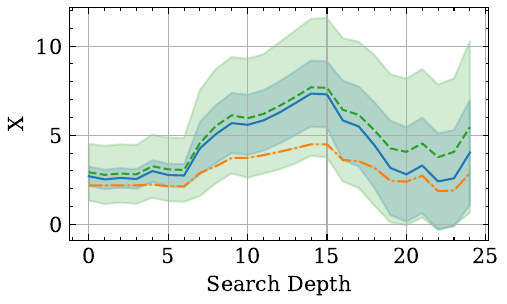}};%
      \end{tikzpicture}%
    }%
  }%
    \subfloat[$\alpha = 0.1$]{%
    \resizebox{0.25\linewidth}{!}{%
      \begin{tikzpicture}%
        \node(a){\includegraphics[width=\linewidth]{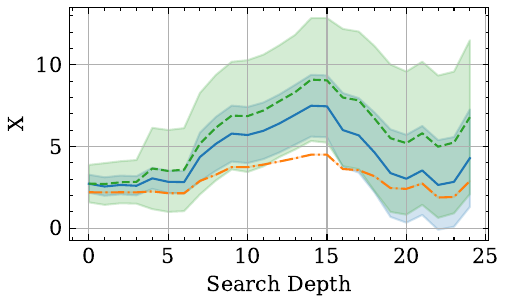}};%
      \end{tikzpicture}%
    }%
  }
    \subfloat[$\alpha = 0.2$]{%
    \resizebox{0.25\linewidth}{!}{%
      \begin{tikzpicture}%
        \node(a){\includegraphics[width=\linewidth]{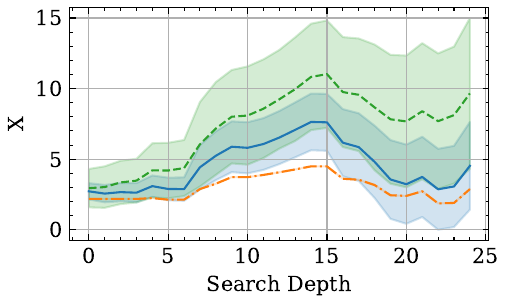}};%
      \end{tikzpicture}%
    }%
  }
    \subfloat[$\alpha = 0.25$]{%
    \resizebox{0.25\linewidth}{!}{%
      \begin{tikzpicture}%
        \node(a){\includegraphics[width=\linewidth]{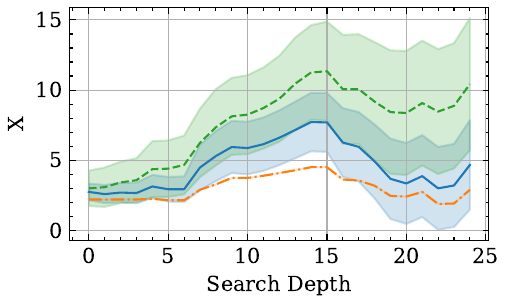}};%
      \end{tikzpicture}%
    }%
  }
  %   \centering
  %   \subfloat[\label{fig:rollouts_stochastic_simple_env}]{%
  %   \resizebox{\linewidth}{!}{%
  %     \begin{tikzpicture}%
  %       \node{\includegraphics[width=\linewidth]{figures/images/visualizing_rollouts/stochastic_trajectories_simple_env_varying_alpha.pdf}};%
  %     \end{tikzpicture}%
  %   }%
  % }%
%   \centering%
%   \subfloat[\label{fig:rollouts_deterministic_pendulum}]{%
%     \resizebox{0.45\linewidth}{!}{%
%       \begin{tikzpicture}%
%         \node(a){\includegraphics[width=\linewidth]{figures/images/visualizing_rollouts/deterministic_trajectories_dubins.pdf}};%
%       \end{tikzpicture}%
%     }%
%   }%
%   \subfloat[\label{fig:rollouts_stochastic_pendulum}]{%
%     \resizebox{0.45\linewidth}{!}{%
%       \begin{tikzpicture}%
%         \node(a){\includegraphics[width=\linewidth]{figures/images/visualizing_rollouts/stochastic_trajectories_dubins.pdf}};%
%       \end{tikzpicture}%
%     }%
%   }%
  \caption{For a fixed action distribution, the state distribution simulated by DSSPD, while planning in Simple Env, is a better approximation of the empirical state distribution than the one due to DSSPD-NV.}\label{fig:viz_rollouts}
\end{figure*}

\myparquestion{Does DiSProD provide a better approximation of the system dynamics than DiSProD-NV?}
As discussed earlier, DiSProD-NV does not account for any uncertainty in the dynamics and ignores the variance terms in the Taylor's expansion while DiSProD uses a 2nd degree Taylor's expansion to approximate the dynamics.
Note that DiSProD-NV does not calculate a distribution but simply propagates the mean.
In addition, its calculated mean is potentially less accurate.
We explore this behaviour in SimpleEnv. For a fixed sequence of action distributions, we compute the state distribution due to DiSProD and DiSProD-NV and compare that to the empirical state distribution visited by the agent when it samples from the same action distribution. In Figure \ref{fig:viz_rollouts} we compare the empirical state distribution of $x$ against the state distributions as computed by DiSProD and DiSProD-NV. 
When $\alpha$ is 0, the approximation using DiSProD overlaps with the empirical distribution for some time before diverging slightly. As the stochasticity of the environment increases (indicated by increasing $\alpha$ values), the state distributions due to DiSProD and DiSProD-NV diverge from the empirical state distribution. But the distribution due to DiSProD is a better approximation than the distribution due to DiSProD-NV.
% Planning horizon

\begin{figure*}[htb]%
  \centering%
    \subfloat{%
        \includegraphics[width=0.4\linewidth]{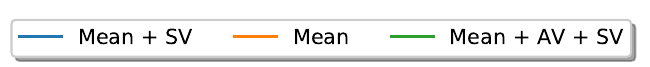}
    }\\
    \setcounter{subfigure}{0}
   \subfloat[CCP\label{fig:ablation_cartpole}]{%
     \resizebox{0.25\linewidth}{!}{%
       \begin{tikzpicture}%
         \node(a){\includegraphics[width=\linewidth]{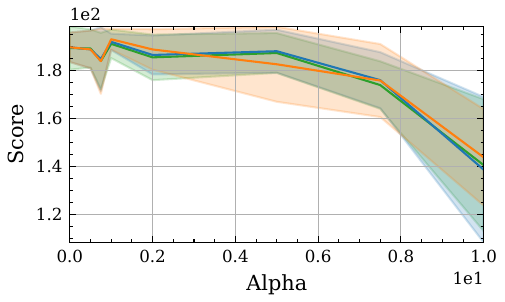}};%
       \end{tikzpicture}%
     }%
   }%
  \subfloat[CMC\label{fig:ablation_mountain_car}]{%
    \resizebox{0.25\linewidth}{!}{%
      \begin{tikzpicture}%
        \node(a){\includegraphics[width=\linewidth]{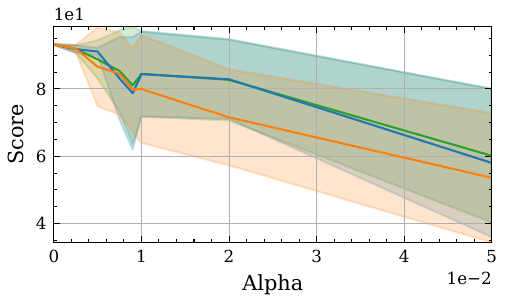}};%
      \end{tikzpicture}%
    }%
  }%
  \subfloat[P\label{fig:ablation_pendulum}]{%
    \resizebox{0.25\linewidth}{!}{%
      \begin{tikzpicture}%
        \node(a){\includegraphics[width=\linewidth]{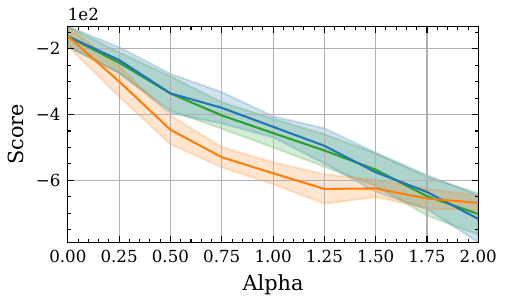}};%
      \end{tikzpicture}%
    }%
  }%
  \subfloat[SE\label{fig:ablation_simple_env}]{%
    \resizebox{0.25\linewidth}{!}{%
      \begin{tikzpicture}%
        \node(a){\includegraphics[width=\linewidth]{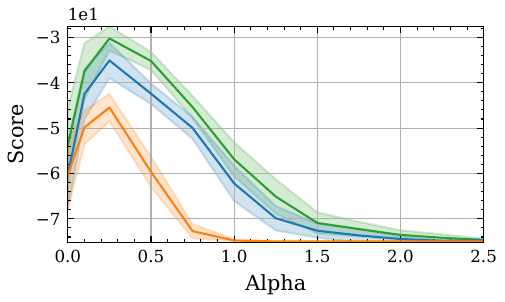}};%
      \end{tikzpicture}%
    }%
  }%
  \caption{Environments used are Continuous Cartpole (CCP), Pendulum (P), Continuous Mountain Car (CMC) and Simple Env (SE). Ignoring the variance terms generally hurts DiSProD, especially with increasing $\alpha$. While action variance does not contribute a lot to the planner's performance in some environments (a, b, c), but in others (d) searching over a stochastic policy yields further improvement.
  }\label{fig:ablation_results}
\end{figure*}

\myparquestion{Do both state and action variance improve planning?}
DiSProD relies on state and action variance terms to propagate distributions. Intuitively, state variance captures the impact of noise on the state variables while having action variance enables DiSProD to search over a stochastic policy. We explore the performance of the three variants of DiSProD discussed in \Cref{sec:analysis_of_ablation_modes}. 
Results are shown in Figure \ref{fig:ablation_cartpole}, \ref{fig:ablation_mountain_car}, \ref{fig:ablation_pendulum}.
We observe that for regions of low stochasticity, all the three modes have similar performance. But with increasing $\alpha$, zeroing out the state variance harms DiSProD. Further, in all of the basic environments, zeroing out action variance but keeping state variance does not harm performance. 
However, as shown in \Cref{fig:ablation_simple_env}, in the Simple Env which has a more complex noise structure (see \Cref{sec:simple_env_details}) action variance leads to further improvement in performance. 

\myparquestion{Detailed results for experiments with modified Dubins Car model}
The detailed results for experiments with the modified Dubins Car model in Gym and TurtleBot are given in \Cref{tab:results-gym-dubins,tab:results-dubins-turtlebot}. These are aggregated in the main paper in \Cref{tab:turtlebotresults} by averaging over the 16 maps.

\begin{table*}
    \centering\scriptsize%
    \begin{tabular}{rrrrrrr}
        \toprule
        \multirow{2}{*}{Map} & \multicolumn{2}{c}{CEM} & \multicolumn{2}{c}{MPPI} & \multicolumn{2}{c}{DiSProD}\\
                             & {\%Success$\uparrow^{100}_{0}$} & {\#Steps$\downarrow$} & {\%Success$\uparrow^{100}_{0}$} & {\#Steps$\downarrow$} & {\%Success$\uparrow^{100}_{0}$} & {\#Steps$\downarrow$}\\\midrule
        no-ob-1&$100.0\pm0.0$&$37.0\pm0.4$&$100.0\pm0.0$&$38.0\pm0.0$&$100.0\pm0.0$&\bm{$35.0\pm1.6$}\\
        no-ob-2&$100.0\pm0.0$&$44.0\pm0.0$&$100.0\pm0.0$&$48.0\pm0.0$&$100.0\pm0.0$&\bm{$42.0\pm1.2$}\\
        no-ob-3&$100.0\pm0.0$&\bm{$63.0\pm0.0$}&$100.0\pm0.0$&$67.0\pm0.8$&$100.0\pm0.0$&\bm{$63.0\pm0.0$}\\
        no-ob-4&$100.0\pm0.0$&\bm{$44.0\pm0.0$}&$100.0\pm0.0$&$49.0\pm0.0$&$100.0\pm0.0$&\bm{$44.0\pm0.0$}\\
        no-ob-5&$100.0\pm0.0$&$39.0\pm2.0$&$100.0\pm0.0$&$37.0\pm0.4$&$100.0\pm0.0$&\bm{$30.0\pm0.0$}\\
        ob-1&$100.0\pm0.0$&\bm{$22.0\pm0.0$}&$100.0\pm0.0$&$29.0\pm0.0$&$100.0\pm0.0$&\bm{$22.0\pm0.8$}\\
        ob-2&$100.0\pm0.0$&$77.0\pm2.71$&$100.0\pm0.0$&$76.0\pm2.86$&$100.0\pm0.0$&\bm{$75.0\pm2.0$}\\
        ob-3&$100.0\pm0.0$&$36.0\pm2.0$&$100.0\pm0.0$&$39.0\pm0.4$&$100.0\pm0.0$&\bm{$31.0\pm1.6$}\\
        ob-4&$100.0\pm0.0$&\bm{$41.0\pm0.0$}&$100.0\pm0.0$&$46.0\pm0.0$&$100.0\pm0.0$&$45.0\pm6.4$\\
        ob-5&$100.0\pm0.0$&\bm{$47.0\pm0.0$} & $100.0\pm0.0$ & \bm{$47.0\pm0.0$} & $100.0\pm0.0$&$49.0\pm0.0$\\
        ob-6&\bm{$100.0\pm0.0$}&\bm{$60.0\pm0.0$}&$80.0\pm40.0$&$133.0\pm133.33$&\bm{$100.0\pm0.0$}&\bm{$60.0\pm0.0$}\\
        ob-7&$100.0\pm0.0$&\bm{$32.0\pm0.0$}&$100.0\pm0.0$&$38.0\pm0.0$&$100.0\pm0.0$&\bm{$31.0\pm3.72$}\\
        ob-8&$100.0\pm0.0$&$44.0\pm0.0$&$100.0\pm0.0$&$48.0\pm2.0$&$100.0\pm0.0$&\bm{$39.0\pm0.8$}\\
        ob-9&$100.0\pm0.0$&\bm{$31.0\pm0.0$}&$100.0\pm0.0$&\bm{$32.0\pm0.0$}&$100.0\pm0.0$&$42.0\pm0.0$\\
        ob-10&$100.0\pm0.0$&$46.0\pm4.0$&$100.0\pm0.0$&$41.0\pm2.4$&$100.0\pm0.0$&\bm{$29.0\pm0.0$}\\
        ob-11&$100.0\pm0.0$&\bm{$45.0\pm0.0$}&$100.0\pm0.0$&$48.0\pm0.4$&$100.0\pm0.0$&$48.0\pm0.0$\\
        cave-mini&$100.0\pm0.0$&$79.0\pm1.6$&$100.0\pm0.0$&$85.0\pm0.0$&$100.0\pm0.0$&\bm{$75.0\pm4.8$}\\
        \bottomrule
    \end{tabular}
    \caption{Success rate and number of steps taken by the planners for each map when run in the Gym environment. The dynamics model used by the planners is an accurate representation of the true dynamics of the system. In most maps, DiSProD takes the least number of steps to reach the goal.}
    \label{tab:results-gym-dubins}
\end{table*}

\begin{table*}
    \centering\scriptsize%
    \begin{tabular}{rrrrrrr}%
        \toprule%
        {Maps} & \multicolumn{2}{c}{CEM} & \multicolumn{2}{c}{MPPI} & \multicolumn{2}{c}{DiSProD}\\
               & {\%Success$\uparrow^{100}_{0}$} & {\#Steps$\downarrow$} & {\%Success$\uparrow^{100}_{0}$} & {\#Steps$\downarrow$} & {\%Success$\uparrow^{100}_{0}$} & {\#Steps$\downarrow$} \\\midrule
        no-ob-1&$100.0\pm0.0$&$83.0\pm12.2$&$100.0\pm0.0$&$134.0\pm90.04$&$100.0\pm0.0$&\bm{$59.0\pm9.56$}\\
        no-ob-2&$100.0\pm0.0$&\bm{$169.0\pm6.95$}&$100.0\pm0.0$&$180.0\pm3.88$&$100.0\pm0.0$&$174.0\pm14.05$\\
        no-ob-3&$100.0\pm0.0$&$163.0\pm7.84$&$100.0\pm0.0$&$195.0\pm7.22$&$100.0\pm0.0$&\bm{$143.0\pm16.77$}\\
        no-ob-4&$100.0\pm0.0$&$116.0\pm3.88$&$100.0\pm0.0$&$121.0\pm2.87$&$100.0\pm0.0$&\bm{$101.0\pm12.67$}\\
        no-ob-5&$100.0\pm0.0$&\bm{$197.0\pm6.08$}&$100.0\pm0.0$&$203.0\pm5.12$&$100.0\pm0.0$&$211.0\pm14.18$\\
        ob-1&$40.0\pm49.0$&$289.0\pm136.44$&\bm{$100.0\pm0.0$}&\bm{$149.0\pm2.1$}&\bm{$100.0\pm0.0$}&$155.0\pm51.79$\\
        ob-2&$100.0\pm0.0$&$89.0\pm2.86$&$100.0\pm0.0$&$120.0\pm2.42$&$100.0\pm0.0$&\bm{$88.0\pm11.62$}\\
        ob-3&$100.0\pm0.0$&\bm{$96.0\pm4.35$}&$100.0\pm0.0$&$136.0\pm4.8$&$100.0\pm0.0$&$97.0\pm5.68$\\
        ob-4&$100.0\pm0.0$&\bm{$132.0\pm10.13$}&$100.0\pm0.0$&$177.0\pm8.52$&$100.0\pm0.0$&$144.0\pm16.81$\\
        ob-5&$100.0\pm0.0$&\bm{$93.0\pm7.91$}&$100.0\pm0.0$&$149.0\pm3.61$&$100.0\pm0.0$&$103.0\pm6.77$\\
        ob-6&$0.0\pm0.0$&$400.0\pm0.0$&$0.0\pm0.0$&$400.0\pm0.0$&\bm{$80.0\pm40.0$}&\bm{$243.0\pm115.9$}\\
        ob-7&$100.0\pm0.0$&\bm{$87.0\pm5.23$}&$100.0\pm0.0$&$127.0\pm4.59$&$100.0\pm0.0$&$94.0\pm4.83$\\
        ob-8&$100.0\pm0.0$&\bm{$133.0\pm1.36$}&$100.0\pm0.0$&$175.0\pm14.43$&$100.0\pm0.0$&$155.0\pm9.4$\\
        ob-9&$100.0\pm0.0$&$112.0\pm7.24$&$100.0\pm0.0$&\bm{$110.0\pm2.68$}&$100.0\pm0.0$&$126.0\pm33.85$\\
        ob-10&$40.0\pm49.0$&$298.0\pm125.42$&\bm{$100.0\pm0.0$}&\bm{$150.0\pm6.09$}&$60.0\pm49.0$&$233.0\pm137.32$\\
        ob-11&\bm{$80.0\pm40.0$}&$239.0\pm80.75$&$60.0\pm49.0$&$275.0\pm102.4$&\bm{$80.0\pm40.0$}&\bm{$230.0\pm88.32$}\\
        cave-mini&$100.0\pm0.0$&\bm{$290.0\pm15.97$}&$0.0\pm0.0$&$400.0\pm0.0$&$100.0\pm0.0$&$332.0\pm20.43$\\
        \bottomrule
    \end{tabular}
    \caption{Success rate and number of steps taken by the planners for each map while controlling Turtlebot. The dynamics model used by the planners in an approximation of the true dynamics of the system. Although in some cases DiSProD takes longer to complete the goal in maps with obstacles, it has a better overall SR as compared to CEM/MPPI.}\label{tab:results-dubins-turtlebot}
\end{table*}

\section{Runtime}
We compare the running time of \disprodSpace and \disprod-NV with CEM and MPPI in the basic Gym environments on a system with 3.5GHz i7 CPU and 32GB of memory. We run each planner for 5 episodes and compute the mean and standard deviation of the time taken for each episode. As shown in the table below, the run time of \disprod-NV is close to both CEM and MPPI, while \disprodSpace is up to $7$ times slower. \disprodSpace computes the diagonal of the Hessian matrix for propagating distribution which is expensive. The partials also add to the cost of the gradient computation while optimizing actions. Both of these computations can be expensive for high-dimensional state spaces.

\vspace{10pt}

% \begin{table}[htbp]
{
  \centering
  \scriptsize
  \begin{tabular}{lcccc}
    \toprule
 Environment & CEM & MPPI & \disprod-NV & \disprod \\ 
 \midrule
 Cart Pole & $3.69 \pm 0.68$ & $3.70 \pm 0.75$ & $4.10 \pm 1.60$ & $22.26 \pm 4.50$ \\
 Pendulum & $3.68 \pm 0.68$ & $3.72 \pm 0.78$ & $4.11 \pm 1.54$ & $18.16 \pm 6.73$\\  
 Mountain Car & $2.04 \pm 0.76$ & $2.24 \pm 0.84$ & $2.66 \pm 1.69$ & $5.70 \pm 2.93$\\
    \bottomrule
  \end{tabular}
 }
% \end{table}
\end{document}